\documentclass[pdflatex,sn-mathphys-num]{sn-jnl}

\usepackage{graphicx}%
\usepackage{multirow}%
\usepackage{amsmath,amssymb,amsfonts}%
\usepackage{amsthm}%
\usepackage[title]{appendix}%
\usepackage{xcolor}%
\usepackage{textcomp}%
\usepackage{manyfoot}%
\usepackage{booktabs}%
\usepackage[linesnumbered, ruled,vlined]{algorithm2e}
\usepackage{listings}%
\usepackage{subcaption}%
\usepackage{url}%

\theoremstyle{thmstyleone}%
\theoremstyle{thmstyletwo}%

\theoremstyle{thmstylethree}%

\newcommand{\vct}[1]{\ensuremath{\mathbf{#1}}}

\newcommand{\eg}{{e.g.}\xspace}

\begin{document}

\title[Article Title]{Over-parameterization and Adversarial Robustness in Neural Networks: An Overview and Empirical Analysis}


\author[1,2]{\fnm{Srishti} \sur{Gupta}}\email{srishti.gupta@uniroma1.it}

\author[3]{\fnm{Zhang} \sur{Chen}}\email{1005811379@mail.nwpu.edu.cn}

\author[4]{\fnm{Luca} \sur{Demetrio}}\email{luca.demetrio@unige.it}

\author[2]{\fnm{Fabio} \sur{Brau}}\email{fabio.brau@unica.it}

\author[3]{\fnm{Xiaoyi} \sur{Feng}}\email{fengxiao@nwpu.edu.cn}

\author[3]{\fnm{Zhaoqiang} 
\sur{Xia}}\email{zxia@nwpu.edu.cn}

\author[4]{\fnm{Antonio Emanuele} \sur{Cinà}}\email{antonio.cina@unige.it}

\author[2]{\fnm{Maura} \sur{Pintor}}\email{maura.pintor@unica.it}

\author[4]{\fnm{Luca} 
\sur{Oneto}}\email{luca.oneto@unige.it}

\author*[2]{\fnm{Ambra} 
\sur{Demontis}}\email{ambra.demontis@unica.it}

\author[2]{\fnm{Battista} 
\sur{Biggio}}\email{battista.biggio@unica.it}

\author[2,4]{\fnm{Fabio} 
\sur{Roli}}\email{fabio.roli@unica.it}


\affil[1]{\orgname{Sapienza University of Rome}, \state{Rome}, \country{Italy}}

\affil[2]{\orgname{University of Cagliari}, \state{Cagliari}, \country{Italy}}

\affil[3]{\orgname{Northwestern Polytechnical University},  \state{Xi'An}, \country{China}}

\affil[4]{\orgname{University of Genoa},  \state{Genoa}, \country{Italy}}


\abstract{Thanks to their extensive capacity, over-parameterized neural networks exhibit superior predictive capabilities and generalization. Nevertheless, having a large parameter space is considered one of the main suspects of the neural network vulnerability to adversarial examples— input samples crafted ad hoc to induce a desired misclassification.  
The relevant literature has presented contradictory claims regarding the robustness of over-parameterized networks. These contradictory findings might be due to the failure of the attack employed to evaluate the robustness of the networks. Previous research has demonstrated that depending on the considered model, the algorithm employed to generate adversarial examples may not function properly, leading to overestimating its robustness. In this work, we empirically study the robustness of over-parameterized networks against adversarial examples. However, unlike the previous works, we also evaluate the considered attack's reliability, employing the tools at the state of the art, to support the results' veracity. Our results show that over-parameterized networks are more robust to adversarial attacks than their under-parameterized counterparts.}




\maketitle

\section{Introduction}
\label{sec1}

Neural networks have achieved outstanding performance over the last decade. This advancement has been possible not only thanks to the availability of large datasets but also due to the large number of parameters employed. The networks that have reported better performance are often over-parameterized~\citep{Belkin_2019}: their  capacity is higher than the one needed to interpolate the training data (to fit them perfectly). 
In modern Machine Learning (ML) practices, rich function classes like neural networks (NN) are highly capable of training models that \textit{generalizes} well on test samples, meaning giving accurate predictions on samples not seen during the training, provided the test data is sampled from the same distribution as the training data~\citep{neyshabur2015search, novak2018sensitivity}. 

\begin{figure*}[t]
    \centering
    \begin{subfigure}[b]{0.45\textwidth}
        \includegraphics[width=\textwidth]{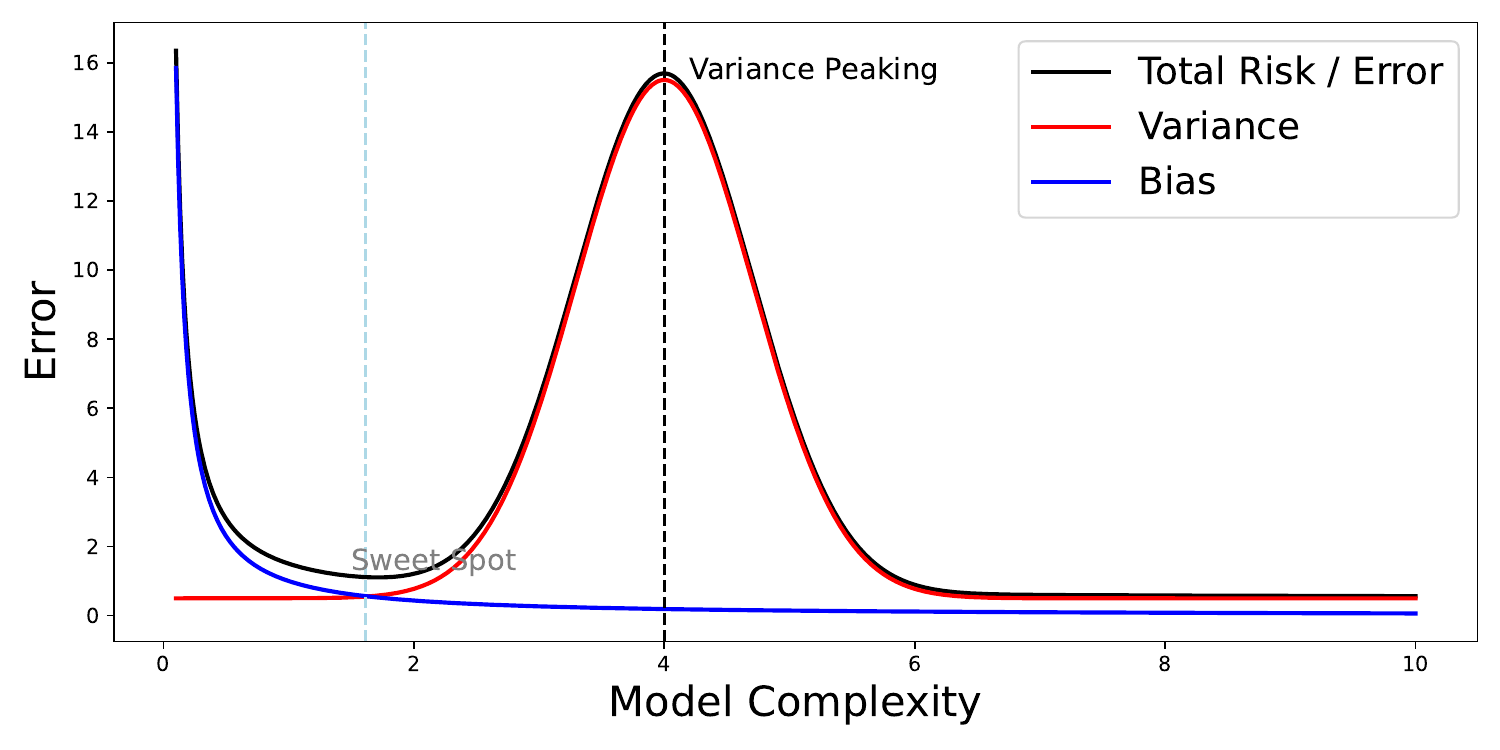}
    \end{subfigure}
    \begin{subfigure}[b]{0.43\textwidth}
        \includegraphics[width=\textwidth]{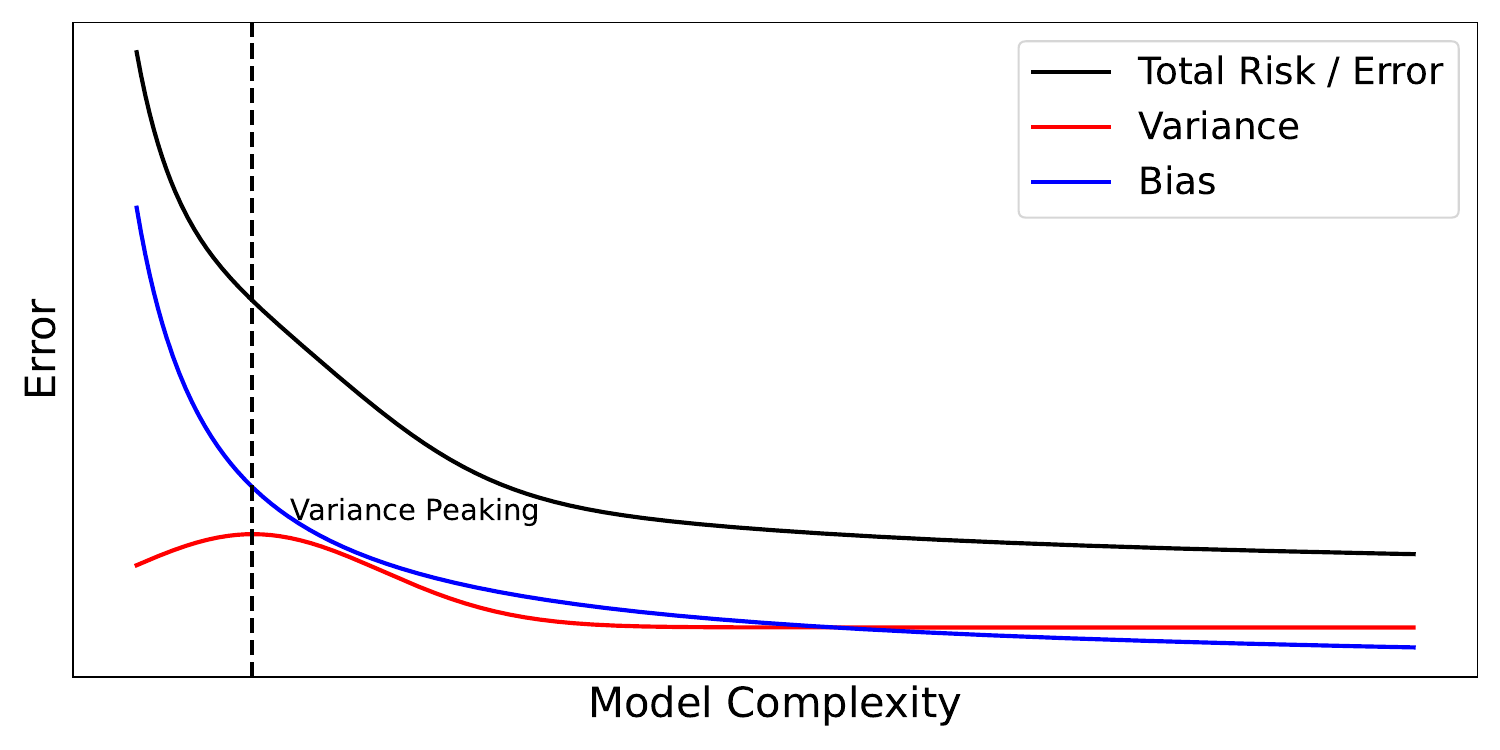}
    \end{subfigure}
    \caption{\textbf{Risk curves.} The bias is represented in blue and the variance is represented in red. Left plot: Double-descent risk curve; Right plot: The risk when, as for deep neural networks, the bias dominates.}
\label{fig:risk_curves}
\end{figure*}

However, this phenomenon is at odds with the classical understanding of bias-variance tradeoff that predicts lower bias in increasingly complex\footnote{The classical bounds in statistical learning theory point to the size of the model in terms of the number of parameters as a measure of complexity.} models at the expense of higher variance, typically supporting parsimonious networks. Such a mismatch in theoretical and empirical understanding has led the community to study the behavior of over-parameterized networks deeply. Belkin et al.'s theory of \textit{double-descent curve} (Figure \ref{fig:risk_curves}) confirms that the variance and, subsequently, the risk of the model decreases as the size of the model goes beyond the interpolation point. 
It is even considered that the lowest generalization error is expected at the infinitely large model. Large neural networks have shown impressive performance at test time across domains, leading to their widescale applicability. However, surprisingly, these models are vulnerable to \textit{adversarial examples}: input samples carefully perturbed to be classified as the attacker desires~\citep{biggio13-ecml,goodfellow15-iclr, nguyen15-cvpr}. Computer vision poses a particularly challenging issue since it has been shown that for the attackers, it is sufficient to apply a tiny perturbation, imperceptible to human eyes, to have a sample misclassified with high confidence ~\citep{szegedy14-iclr, moosavi16-deepfool}. For security-sensitive applications such as autonomous cars and face recognition, it is imperative that the models are robust against adversarial examples while performing well on natural samples. From a classical understanding of the optimal size of the neural network, most neural networks, to minimize the empirical risk, were biased to choose the smallest possible model that fits the training data well, i.e., stay on the left side of the double descent curve.
Hence, most adversarial evaluations are performed on these under-parameterized networks. Recently, given the massive performance hike of over-parameterized models and their generalizable properties, their robustness to adversarial examples has been investigated in multiple works. 
However, the claims about the security of these models are contradictory, and the relationship of over-parameterized models with adversarial robustness is unclear. Recent empirical studies showed their near-optimal results even when the neural network and kernel machines trained to interpolation are corrupted with noise ~\citep{zhang2017understanding, belkin2018understand}. 
Adversarial examples, distinct from random noise, are carefully crafted samples designed to intentionally target the vulnerability of the model. One of the major reasons for these contradictory results may be the unreliable evaluation of adversarial robustness. In various recent works where a robustness evaluation was performed, adversarial attacks were found to have failed~\citep{tramer2020adaptive}, i.e., the attacks generated were ineffective and hence useless for evaluation, giving a false sense of security ~\citep{popovic2022gradient}. However, the articles assessing the robustness of overparameterized networks with respect to their non-overparameterized counterpart have never investigated the effectiveness of the attack employed to perform the robustness evaluation. We evaluate the adversarial robustness of models trained across a spectrum of capacities against Projected Gradient Descent (PGD) attacks. To be consistent with best practices and to reliably evaluate robustness, we test the models using the AutoAttack framework. To avoid known attack failures, we also used tools to verify the attack's effectiveness. The empirical results from the experiments suggest that highly parameterized models are more robust than under-parameterized models. We discuss the background and the notations in \autoref{sec: background}, the methodology adopted to conduct experiments in \autoref{sec: study_design}, and experimental setup and results in \autoref{sec: experiments}. We conclude by discussing the major takeaways and future research directions in \autoref{sec: results}.\\
The code is available at \url{https://github.com/pralab/overparam-adv}

\section{Over-parameterization and Adversarial Robustness}
\label{sec: background}

In this section, we discuss the need and application of over-parameterized models, their vulnerability to adversarial attacks, and what the existing literature suggests on the robustness of these large models.

\subsection{Over-parameterization and Double-descent Curve}

When designing a model architecture to solve an ML problem, from classical understanding, we choose a model that is rich enough to capture the underlying feature space without fitting too closely with the training data (or overfitting) and simple enough to avoid not being able to learn patterns in the data (or underfitting). The balance between underfitting and overfitting is widely known as \textit{bias-variance tradeoff} plotted on a risk capacity curve as in Figure \ref{fig:risk_curves}. From this understanding, we would want to find a "sweet spot" where the model has enough capacity to perform well on test data while keeping the training error low. Having a large model with zero training error is conventionally avoided, as stated by Trevor Hastie in~\citep{hastie_09_elements-of.statistical-learning}, "a model with zero training error is overfit to the training data and will typically generalize poorly” which is still a widely accepted notion. However, large NN are highly expressive and can efficiently optimize many parameters to fit the training data. Large number of parameters of the neural network, traditionally considered a case of overfitting, are better suited for generalization ~\citep{neal2019modern}. Further, Belkin et al.'s ~\citep{Belkin_2019} who coined the term \textit{double-descent curve} (left plot of Figure \ref{fig:risk_curves}) suggested that when training a model on the increasing function capacity (referred to the size of network) until the point of zero training error/interpolation, typically have high risk and poor generalization error. 
Although, as explained in \cite{yang2020rethinking} it is not always a double descent curve, especially for deep neural networks, for which the bias usually dominates with respect to the variance and thus the risk is monotonically decreasing (right plot of Figure \ref{fig:risk_curves}), unless there is noise in the training labels. 
Training beyond the interpolation point (in over-parameterized regime), it is observed that the variance of the model decreases, which leads to the overall decrease in the risk, giving improved generalization performance. 
Further work by  Nakkiran et al.~\citep{nakkiran2019deep} showed that this generalization trend is not just a function of model capacity but also a function of the number of epochs in training. In some literature, it is also argued that good generalization performance is due to the inductive bias of gradient-based algorithms, \eg ~\citep{bartlett2021deep, hastie2020surprises}.

\subsection{Adversarial Robustness}

Despite the unprecedented success and wide adoption of large NN in various domains and their outstanding performances, these models are known to be highly vulnerable to carefully crafted perturbations of the input. For example, consider an input image \(\mathbf{x}\), a vector of pixel information, there can exists visually similar images \(\mathbf{x'}\) with different classifications. Such manipulated input samples are called adversarial examples. 
These are the results of an optimization problem formulated as:
\begin{equation}
    \textrm{max}_{\mathbf{x'}}\quad l(\mathbf{x'}, y, \theta),
\end{equation}
\begin{equation}
    \textrm{s.t.}\quad \|\mathbf{x'} - \mathbf{x}\|_p \leq \epsilon,
\end{equation}
\begin{equation}   
        \mathbf{x}_{lb} \preceq \mathbf{x'} \preceq \mathbf{x}_{ub}
        \label{eq:box_constraint}
\end{equation}
where \(l(.)\) is a loss function, \(\|\mathbf{x'}-\mathbf{x}\|_p\) is the \(l_{p}\) norm constraint bounded by maximum perturbation \(\epsilon\). Eq.\ref{eq:box_constraint} is the box-constraint where \(\mathbf{u}\preceq\mathbf{v}\) means, each element in \(\mathbf{u}\) has to be smaller than or equal to the corresponding element in \(\mathbf{v}\) to bound the values of attack sample \(\mathbf{x'}\).

To measure the adversarial robustness of the models, we use \textit{Security Evaluation Curves (SEC)}. SEC evaluates the ML models for the maximum error obtained against an increasingly admissible level of perturbations of the input images. As we will see later in the experiments, SEC is used to evaluate all the models considered. Models with a smaller area under the curve suggest higher robustness and vice versa.

\subsection{Adversarial Robustness of Over-parameterized Network}
\label{sec: model_capacity}

While a considerable number of works \citep{Wu21-neurips,rice20-ICMLa,Huang21-neurips,Gao-NEURIPS2019,Hassani24} assess the robustness of adversarially trained networks, which is beyond the scope of this article, the findings of related works suggest that overparameterization decreases robustness. In the seminal work by Fu et al. \citep{Fu93-ICNN}, the authors claim that decreasing the number of nodes reduces the sensitivity to small perturbations of the input samples. Sagawa et al.~\citep{sagawa2020investigation} suggested that the over-parameterization can cause the exacerbation of \textit{spurious correlation}, i.e., the relation between the biased features of input like background, texture, or unimportant pixels with the labels. However, in \citep{madry201-iclr}, the authors showed that wider neural networks tend to be more robust, as assessed by the PGD attack. The results in this paper were also replicated in \citep{borji2022-arxiv} and are supported by \citep{LiNeurips2022WhyRobust}, where the authors claim that achieving robustness may require exponentially more parameters than achieving accuracy on clean test data. The contradiction present in the literature has been highlighted in \citep{NEURIPS2022_ea5a63f7}, which 
provides a theoretical analysis that only encompasses two-layer neural networks, and in \citep{10048547}, which, with an analysis limited to linear models, shows that over-parameterization may be either a source of additional robustness or brittleness. 
The analysis performed on deep neural networks by the works affirming that over-parameterization increases robustness \citep{madry201-iclr, borji2022-arxiv} is empirical and based on the PGD attack, and they have not employed any tool to check the reliability of the obtained results. Therefore, the attack might have failed \citep{carlini17-aisec,pintor2022indicators}, providing unreliable results that have caused a false sense of robustness for the overparameterized networks. Because gradient-based attacks can fail in many cases \citep{carlini17-aisec,pintor2022indicators}, e.g., if the employed hyperparameters are incorrect or the decision surface is not smooth enough, they can not achieve convergence, providing a suboptimal point that may not be able to evade the classifier. The goal of this work is to verify, employing the tools available at the state of the art, if a known failure is the reason that led to the abovementioned contradiction between the state-of-the-art works. Analyzing the reliability of the empirical analysis to assess the robustness of overparametrized networks, we have demonstrated that it does not present any failure known to date; thus, the higher robustness of overparametrized networks is not due to known experimental problems.

\section{Study Design}
\label{sec: study_design}

In this section, we discuss the assumptions and methodology considered to conduct the experiments. For the evaluation of the over-parameterized models, we consider a simple yet widely used case of supervised learning. In this work, range of models (with different sizes) are trained on training data and then evaluated with benign and adversarial setting at test time. Before diving deep into the considerations of the threat model used in this paper and their evaluations, we first introduce the notations used throughout the paper.

\subsection{Notations}

We consider a data space [\(\mathcal{X}\), \(\mathcal{Y}\)] where \(\mathcal{X}\) is a \(d\)-dimensional feature space of input samples and \(\mathcal{Y}\) is a label space with c-classes such that \(\mathcal{Y}\in[c]\). 
The training data is represented as \(\mathcal{D} = (\mathbf{x}_i, y_i)_{i=1}^n\) where \(n\) is the size of the training set. We use \(L(\mathcal{D}, \theta)\) to represent \textit{loss} incurred by classifier \(f\in\mathcal{H}: \mathcal{X}\rightarrow \mathcal{Y}\) where \(\mathcal{H}\) is a class of function.
Typically the loss is computed by averaging the sample-wise loss function \(l(f(\mathbf{x}), y)\) over the training set, i.e., \(\frac{1}{n}\sum_{i=1}^n l(f(\vct x_i), y_i)\).
We assume that the classifier \(f\) is learned by minimizing the loss function and best function class \(f^*\) is selected using \textit{empirical risk minimization}:

\[f^* \in \arg\ \min_{f\in\mathcal{H}} \sum_i l(f(\textbf{x}_i), y_i) \, .\]

\subsection{Attack Approach}

The attack approaches considered in our experiments have the following threat model: The attacker's main goal is to cause an integrity violation to evade detection without compromising normal operation using \textit{indiscriminate} attack whose goal is to have the test instance misclassified as any class except the true one. The attacker's capability is limited to influencing data samples at test time to evade the trained model, making it an \textit{evasion} attack. We consider both \textit{white-box} setting, where the attacker has complete access to both model and data distribution, and the \textit{black-box} setting, where the attacker has no access to either data or distribution. To evaluate the security of over-parameterized models, we use the widely used white-box attack approach, which generates attacks on the known trained model parameters. PGD is a widely used evasion attack algorithm to evaluate the robustness of a model and is a good starting point for security evaluation. After evaluating on PGD, as explained in later sections, we perform experiments a on sophisticated diversified ensemble attack: AutoAttack.

\subsubsection{Projected Gradient Descent (PGD)}
Proposed by Madry et al.~\citep{madry18-iclrFhar}, PGD is an iterative optimization-based attack widely used for the evaluation of the models' robustness. 
This algorithm leverages Gradient Descent to compute a perturbation $\vct \delta$ on a sample $\vct x$ originally labeled as $y$, that maximizes the output of a loss function $\ell$ of choice, computed on the target model $f$.
%
%
%
As illustrated in \autoref{algo:attack_algo}, at each iteration (line \ref{line:for}), PGD computed the gradient of the loss w.r.t the perturbation $\vct \delta$ (line \ref{line:grad_update}), which is added to the previously-computed perturbation to maximize the error.
The gradient is post-processed by considering only its sign, modulated by the step size $\alpha$.
Then, the perturbation $\vct \delta$ is projected with $\prod$ inside the feasible domain $\Gamma$ (line \ref{line:constraints}).
Such a domain defines both (i) the capability of the attacker, formalized as a bound $\epsilon$ on the \textit{p}-norm of the perturbation itself resulting in $\parallel \vct \delta \parallel_p \leq \epsilon$ and (ii) the domain constraints needed to create valid and uncorrupted adversarial examples.
At the end of the iterations, the perturbation that mostly maximizes the loss is added to the original sample to produce the adversarial example (line \ref{line:return}).

\begin{algorithm}[t]
    \SetKwInOut{Input}{Input}
    \SetKwInOut{Output}{Output}
    \Input{$\vct x, y$, the initial sample and its label; $f$, the target model; $\ell$ the loss function; $\prod_{\Gamma}$, the projection operator enforcing the constraints of the feasible domain $\Gamma$; $n$, the number of iterations; $\alpha$, the step size}
    \Output{$\vct x^{\star}$, the solution found by the algorithm}

    $\vct \delta_0 \leftarrow \vct 0$

    \For{$i \in [1, n]$}{
        \label{line:for}
        $\vct \delta_i \leftarrow \alpha \, sign(\nabla_{\vct \delta_{i}}\, \ell(f(\vct x + \vct \delta_i), y))$\label{line:grad_update}

        $\vct \delta_{i} \leftarrow \prod_{\Gamma}(\vct x, \vct \delta_i)$\label{line:constraints}
    }
    \textbf{return} $\vct x^{\star} \leftarrow$ \vct x + \texttt{best}$(\vct \delta_0, ..., \vct \delta_{n})$\label{line:return}
\caption{Projected Gradient Descent.}
\label{algo:attack_algo}
\end{algorithm}

\subsubsection{AutoAttack}
\label{sec: AA}

AutoAttack is an ensemble of different typologies of attacks~\citep{croce2020reliable}. 
It leverages three white-box and one black-box attack. The three white-box attacks include Automated PGD with Carlini-Wagner loss \(APGD_{CW}\), Automated PGD with Difference of Logits Ratio loss \(APGD_{DLR}\), targeted Fast Adaptive Boundary (FAB) attack, and Square Attacks as the black-box attack ~\citep{andriushchenko2020square}. 
The Automated PGD (APGD) is the standard PGD attack with a fixed number of iterations and adaptive step size. It uses Carlini-Wagner Loss and Difference of Logits Ratio loss instead of cross-entropy loss. \(APGD_{CW}\) is an untargeted attack whereas \(APGD_{DLR}\) is a targeted attack. Namely, it computes the adversarial example ad-hoc to have it misclassified as a specific class. If it does not succeed, it considers a different class as the target class. The third parameter-free white-box attack is the targeted version of the Fast Adaptive Boundary (FAB) attack ~\citep{croce2020minimally}. Finally, Square Attack is an untargeted black-box attack based on random search and is not affected by gradient masking. Gradient masking is a common adversarial defense~\citep{athalye18} against gradient-based attack algorithms that makes the gradient usually computed by these attacks uninformative, which makes the attacker unable to compute an effective adversarial example. However, thanks to the transferability property of adversarial examples \citep{demontis19-usenix}, these defense techniques can be easily broken by computing an attack against a model owned by the attacker (surrogate). These four attacks are packaged in a single module that serves as a diversified attack and helps evaluate the reliability of the results we obtained from our primary PGD experiments.

\subsection{Evaluation Procedure}

\label{sec: eval_proc}
To evaluate the security of neural networks, we take into account different neural network architectures. These architectures are then scaled with different numbers of hidden units in their layers as done in~\citep{neyshabur2018understanding}. We consider a model architecture, expanded to ten models ranging from under-parameterized to over-parameterized regime \(\theta\in\{\theta_1, \theta_2, .., \theta_m\}\). The models are first trained on the training data. Then, for each model $\theta_i$, its base performance is evaluated using benign test data, i.e., where there is no perturbation \({\epsilon_0}\). Consequently, we generate the adversarial examples ad-hoc against the considered model $\theta_i$ for varying degrees of attack budgets \(\epsilon\in \{\epsilon_1, \epsilon_2 .., \epsilon_k\}\) and for each perturbation budget we compute the \textit{Error Rate} (namely the percentage of misclassified samples) of $\theta_i$ on them. These error rates are then used to generate the SEC curve that shows how the error rate of the model $\theta_i$ increases for a growing perturbation budget of the attacker. 

 

\subsection{How Reliable Are These Evaluations?}

In an attempt to verify the generalizability of the models, different works carried out analysis under different scenarios, ex., real-world environment, adversarial environment, noisy environment, etc. For safety- and security-sensitive tasks, robustness under an adversarial environment is of utmost importance. Various earlier works ~\citep{athalye18, carlini2017adversarial, pintor2022indicators} have shown that constructing effective adversarial defense is harder than expected. Many works that propose new defenses, to evaluate the system security, use suboptimal attack methods, often leading to a false sense of security. Even the use of simple gradient-based attacks like PGD, as observed in ~\citep{mosbach2019logit, croce2019scaling} can fail under certain parameter choices leading to over-estimation of robustness. As explained in the previous section, the related works (e.g., \citep{madry18-iclrFhar, borji2022-arxiv}), have employed PGD to compare the robustness of networks with different sizes, but they have not verified if the attack was failing due to a known problem. Therefore, it is necessary to verify the reliability of the security evaluations. In this work, we verified the results obtained with the PGD attack using Indicators of Attack Failures (IoAF) presented by Pintor et. al ~\citep{pintor2022indicators}, which encompasses all the failures known to date. We also used AutoAttack as an additional ensemble of attacks to measure the reliability of the results shown in \autoref{sec: experiments}. With over decade-long research on devising new attacks and failures to study the robustness of models, many recent works have shown that the attacks may fail to generate viable adversarial samples, for example, depending on the choice of loss function \(L\) and the target model \(\theta\), some loss functions may exhibit \textit{obfuscated gradients}~\citep{Carlini2017evalrobust, athalye18} which prevents gradient-based attacks from generating an optimal attack even when they exist in the feasible domain. Another known issue in constructing attacks is \textit{attack-optimization failures} which are observed at the time of running gradient-based attacks ~\citep{tramer2020adaptive, athalye18-iclr}, which may be either due to usage of suboptimal parameters or inappropriate implementation. Therefore, we use IoAF~\citep{pintor2022indicators}, which is an automated testing package aimed at detecting \textit{known} failures within the optimization process of gradient-based attacks. We use IoAF to make sure the attacks generated on over-parameterized models do not present any failure known to date. Most model evaluations rely on the results of PGD or its weaker version: FGSM~\citep{goodfellow15-iclr}. Models evaluated on single or weak attacks can run into the problem of suboptimal solution found on a local minima that maybe far from the global optimum. Therefore, for reliable evaluation of the over-parameterized models, we run a diverse attack to increase the probability to find better approximation of the global optimim. We use AutoAttack to add diversity in the evaluation. Their parameter-free, computationally cheap and user-independent parameters make it a useful tool for measuring the reliability of the security evaluations of the model (explained in detail in \autoref{sec: AA}).

\begin{figure*}[!htb]	
	\centering
 
    \begin{subfigure}[b]{0.3\textwidth}
	\includegraphics[width=\textwidth]{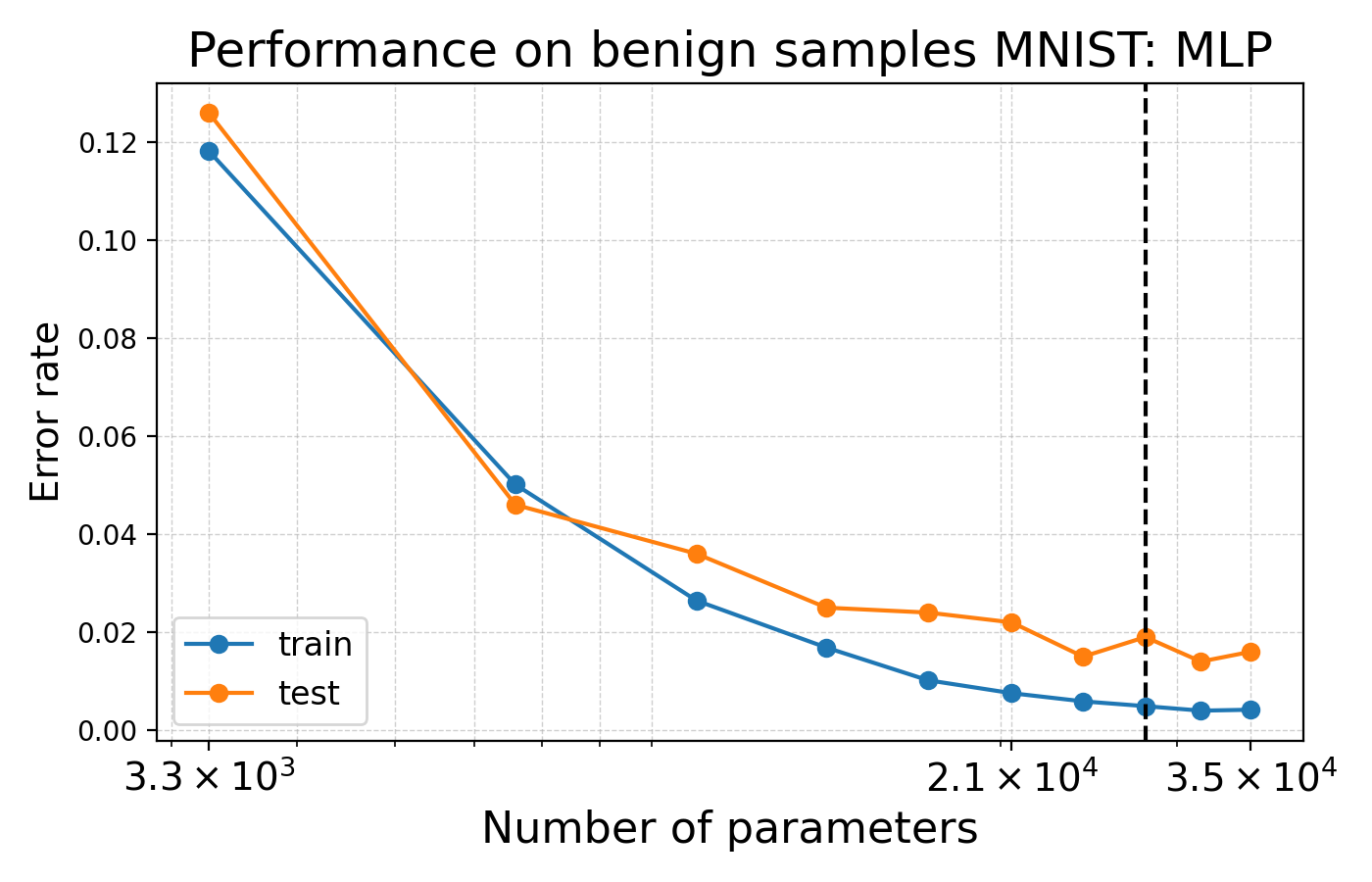}
    \caption{}
    \end{subfigure}
    \begin{subfigure}[b]{0.3\textwidth}
    \includegraphics[width=\textwidth]{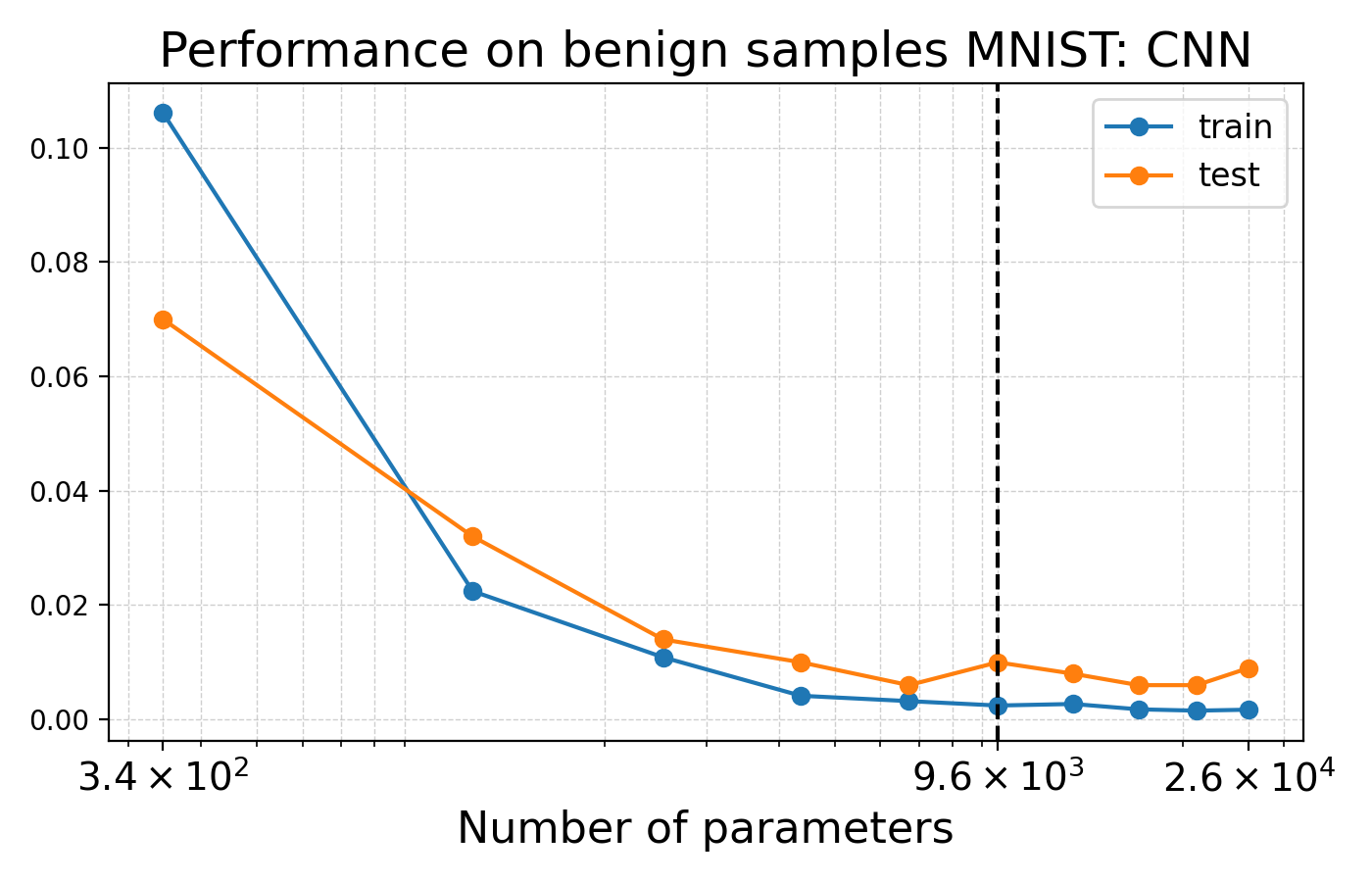}
    \caption{}
    \end{subfigure}
    \begin{subfigure}[b]{0.3\textwidth}
	\includegraphics[width=\textwidth]{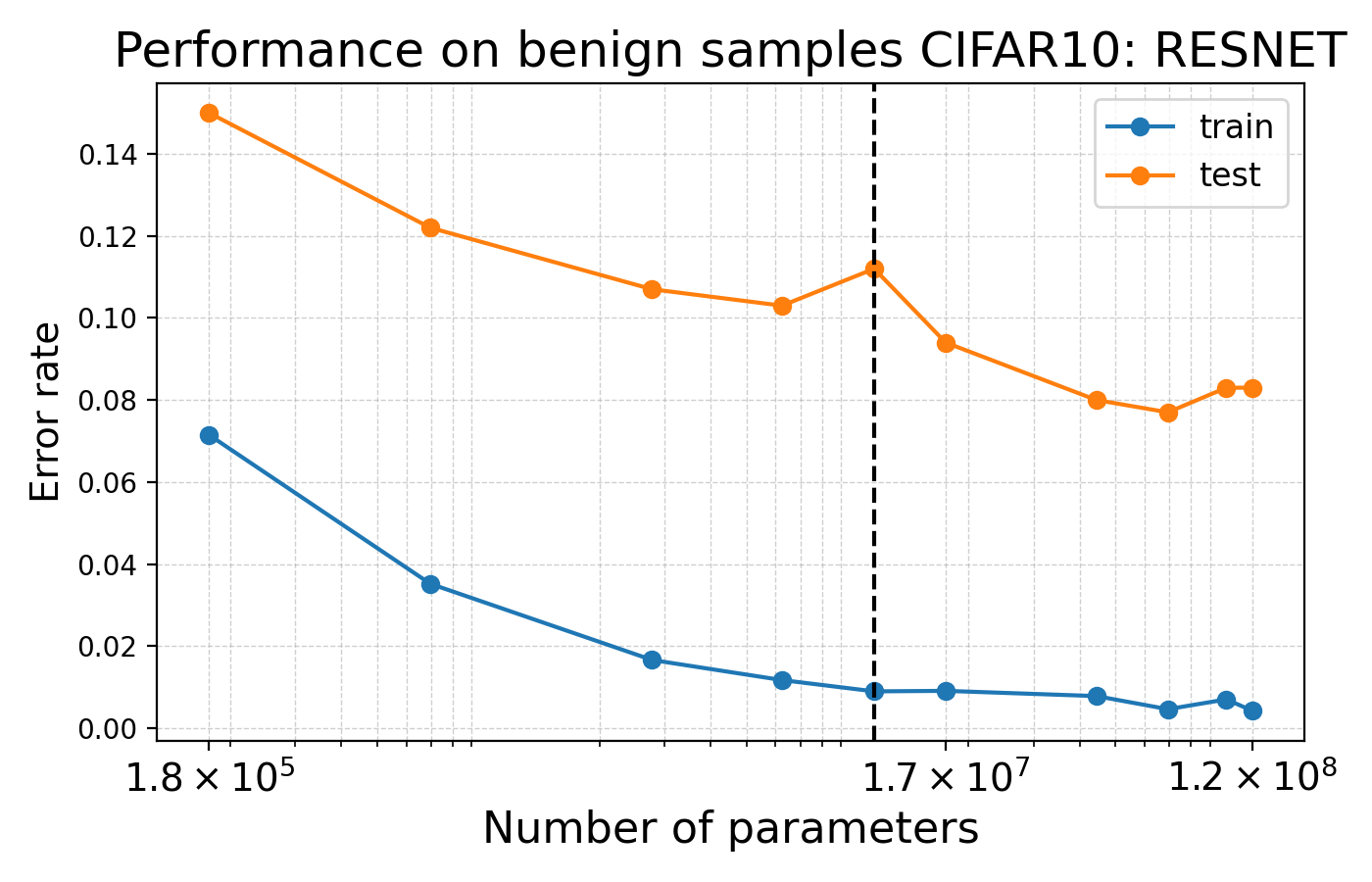}	
    \caption{}
    \end{subfigure}
    
    \caption{\textbf{Base performance.} Evaluation of models on clean samples
    with respect to increasing model size: (a) MLP trained on MNIST, (b) CNN trained on MNIST, and (c) ResNet model trained on CIFAR10. }
    \label{fig: base_performance}
\end{figure*}

\section{Experiments} \label{sec: experiments}

We perform experiments with base neural network architectures that are expanded width-wise by adding neurons to its layers to move the model to the over-parameterized regime.

\subsection{Experimental Setup}
\label{sec: exp_setup}

\paragraph{Datasets.} We conduct experiments on two widely used benchmark datasets: MNIST~\citep{LeCun2005TheMD} and CIFAR-10~\citep{Krizhevsky2009LearningML}.
MNIST consists of 70,000 grayscale images of handwritten digits (0–9), each of spatial resolution 28×28. The dataset is partitioned into 60,000 training samples and 10,000 test samples. CIFAR-10 consists of 60,000 RGB images of size 32×32, evenly distributed across ten object categories. The standard split contains 50,000 training samples and 10,000 test samples.
For both datasets, we further partition the original training set into an 80:20 train–validation split. Models are trained exclusively on the training portion, while the validation set is used for model selection and early checkpointing.
For evaluation, including both clean and adversarial robustness assessment, we use a fixed subset of 1,000 test samples. This choice is motivated by computational considerations: the cost of adversarial attacks scales linearly with the number of evaluation samples and significantly with model capacity.

\paragraph{Models.} We consider three model families:
a) Multi-Layer Perceptron (MLP), b)Convolutional Neural Network (CNN), and c) Residual Network (ResNet).
Each architecture is systematically scaled across 10 capacity levels, yielding 10 distinct model sizes per architecture. Scaling is achieved by increasing the width (number of hidden units or channels) while preserving architectural depth and structural design principles. This controlled scaling enables us to isolate the effect of over-parameterization on both clean and adversarial performance.
 
Formally, let $x \in \mathcal{X} \subset \mathbb{R}^d$ denote the input and $y \in \mathcal{Y} \subset \mathbb{R}^c$ denote the output space. Each model defines a function $f_\theta : \mathbb{R}^d \rightarrow \mathbb{R}^c$ parameterized by $\theta$, where the total number of parameters varies across capacity levels. For instance, in a two-layered feedforward MLP network:
\begin{equation}
    f(x) = h^{(2)}(h^{(1)}(x))
\end{equation}
where $h^{(1)} \in \mathbb{R}^{m\times d}$ and $h^{(2)} \in \mathbb{R}^{c\times m}$, and $m$ controls the model capacity. Similarly, for CNNs and ResNets, capacity scaling increases the size of the convolutional channels while preserving the architectural topology. Across all experiments, architectural hyperparameters other than the width are held constant to ensure that the performance differences arise solely from parameter scaling.

\paragraph{Training Procedure}

Before training, we do a hyperparameter tuning using a validation split. The selected optimization configurations are: a) for MLP and ResNet, we use Stochastic Gradient Descent (SGD) as the optimizer with 0.01 learning rate and 5e-4 weight decay, and b) for CNN, we use ADAM as the optimizer with 1e-3 as the learning rate and 1e-4 as the weight decay.
The models are trained for 1000 epochs. The best-performing checkpoint, defined as the model with the highest validation accuracy, is retained for evaluation. All models are trained using cross-entropy loss. The same training protocol is applied across all capacity levels to ensure comparability.

\paragraph{Evaluation.}
Evaluation is performed in a controlled setting here clean and adversarial performances are computed. \textbf{Benign Performance}: Performance is computed using the saved checkpoints (across all ten capacity levels per architecture) for unperturbed samples. \textbf{Adversarial Performance} Adversarial robustness is evaluated using two complementary attack frameworks: a) Projected Gradient Descent (PGD-L2) implemented via the SecML library and b) AutoAttack, a parameter-free ensemble attack. For each trained checkpoint, adversarial attacks are performed over a range of perturbation budgets $\epsilon$. Robust accuracy is computed as the fraction of correctly classified adversarial examples at each perturbation level.

Experimentally, the evaluation protocol is as follows: a) Load the checkpoint for a given capacity, b) Evaluate clean accuracy on a 1000-sample test subset, c) For each perturbation budget $\epsilon$, generate adversarial examples and compute robust accuracy. These steps are repeated for all architecture-dataset pairs.

\subsection{Experimental Results}

Despite the controversy over the generalizability of over-parameterized networks, we reliably observe greater robustness in largely-parameterized models than in smaller networks.

\subsubsection{Base Performance}
\label{sec: base_perf}

From \autoref{fig: base_performance}\, we see the base performance, namely the performances on the clean (unperturbed) test data, of the three configurations considered in this work: CNN trained on MNIST, MLP trained on MNIST, and Resnet18 trained on CIFAR10, respectively. We can observe that:
\begin{itemize}
    \item The classifier's performance changes when we increase the number of network parameters;
    \item None of the performance on the test data worsens as would have been expected from the classical theory of overfitting for largely parameterized models. Instead, the error rates continue to decrease for over-parameterized networks, supporting the generalizability claims.
\end{itemize}
These are well-known results, but they serve to us as a baseline. It's worth noting that the focus of the work is not on acquiring the best possible classification performance on the base model but instead how the robustness of the model varies as we increase the size of the model.

\subsubsection{Model Robustness}
\label{sec: model_robustness}

\begin{figure*}[!htb]
    \centering
    \begin{subfigure}[b]{0.3\textwidth}
    \centering
    \includegraphics[width=\textwidth]{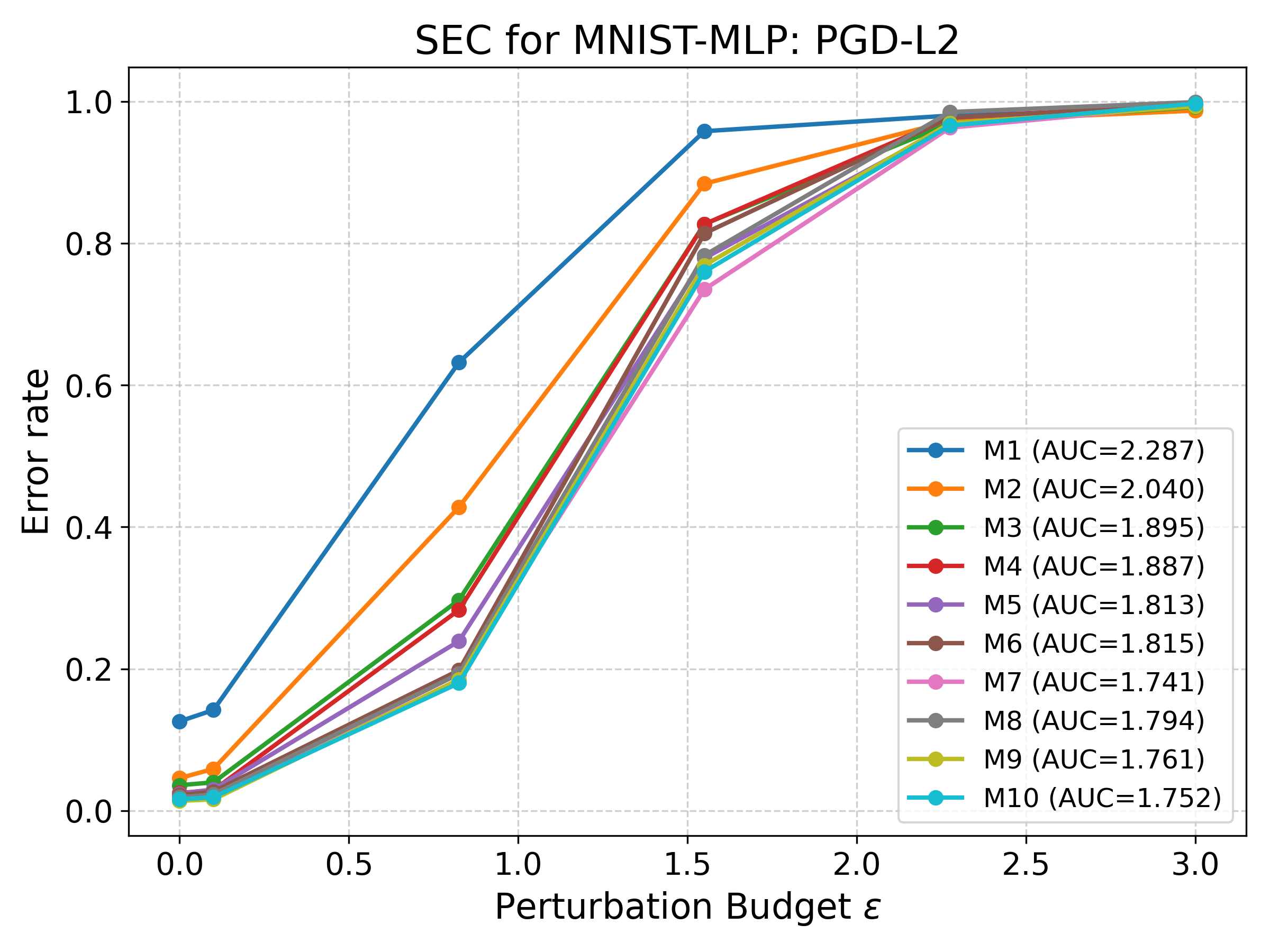}
    \caption{}
    \end{subfigure}
    \begin{subfigure}[b]{0.3\textwidth}
    \centering
    \includegraphics[width=\textwidth]{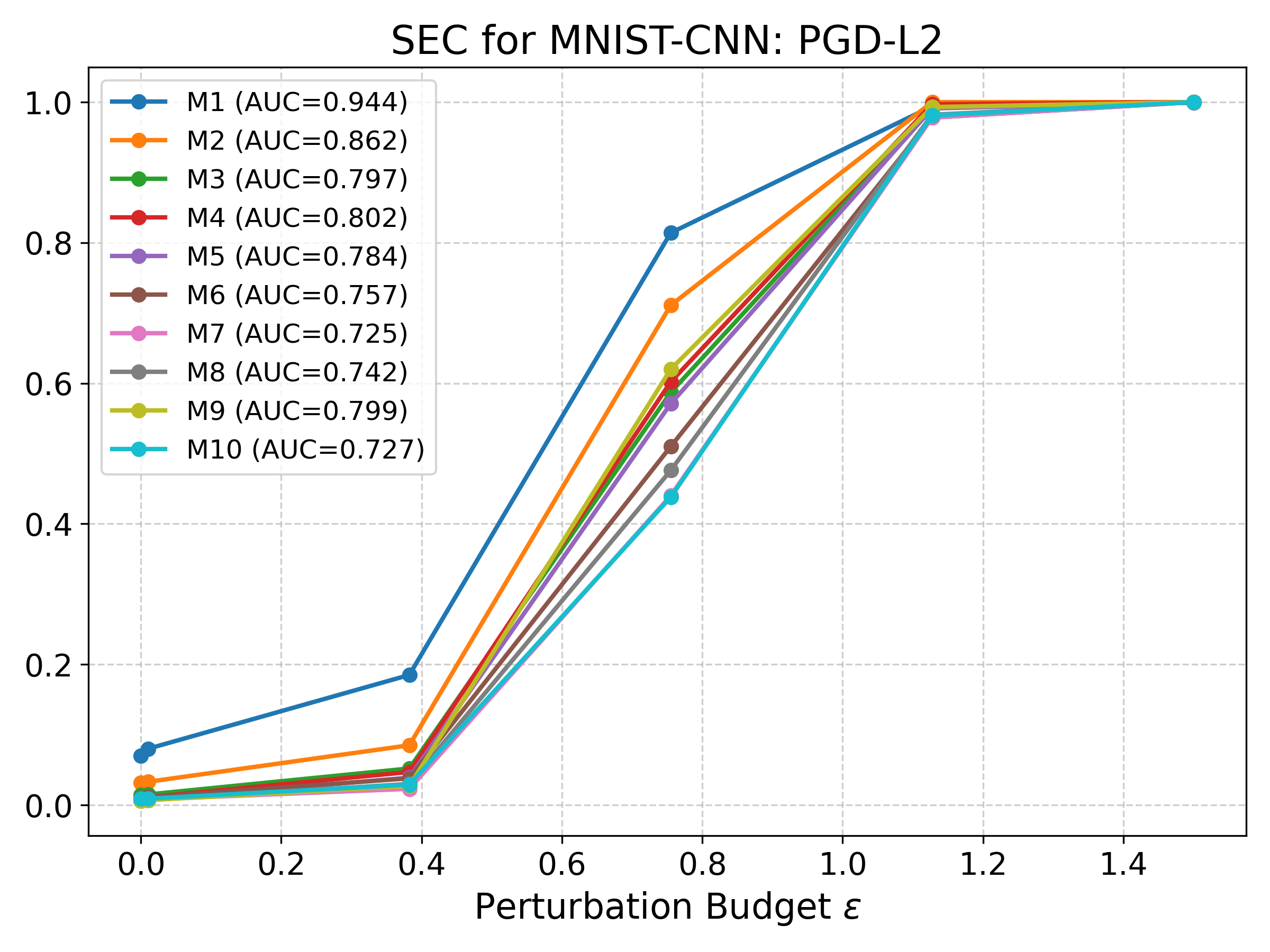}
    \caption{}
    \end{subfigure}
    \begin{subfigure}[b]{0.3\textwidth}
    \centering
    \includegraphics[width=\textwidth]{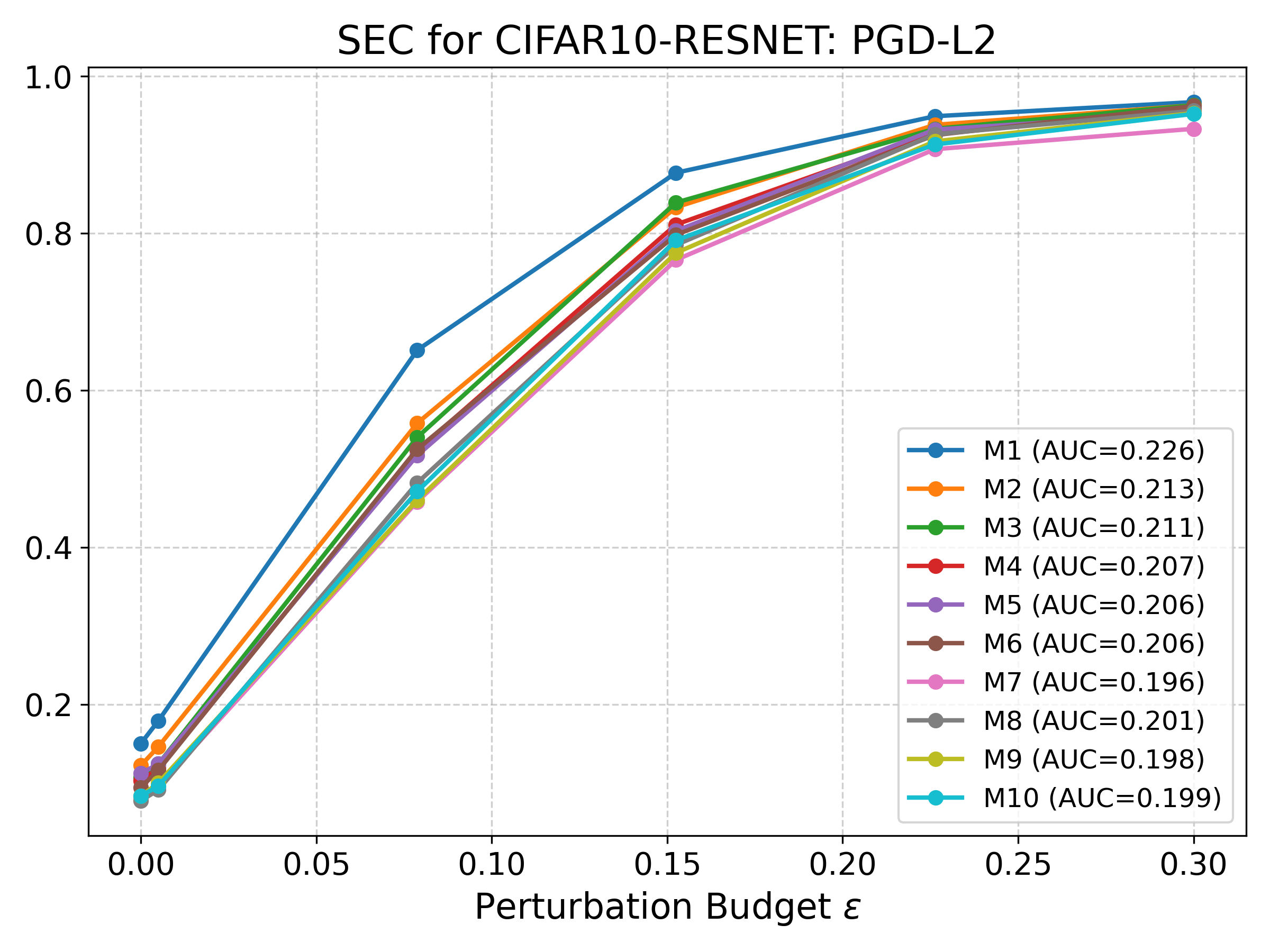}
    \caption{}
    \end{subfigure}
    \hfill
    
    \caption{\textbf{SEC Plots.} Evaluation of varying-sized models under \textbf{PGD $l_2$} attack: (a) MLP trained on MNIST, (b) CNN trained on MNIST, and (c) ResNet trained on CIFAR10. Each curve represents the security evaluation curve with the area under the curve specified in the legend.}
    \label{fig:performance_pgdl2}
\end{figure*}

\autoref{fig:performance_pgdl2} shows the security evaluation curves (SEC) for networks attacked with a PGD attack with an \(l_2\)-norm constraint across an increasing number of parameters and various degrees of perturbation. While it is understood that the error rate would increase with increasing degrees of perturbations, it is interesting to note that \textit{error rates for the over-parameterized model are significantly lower than those of the models with fewer parameters}. We see similar trends for all dataset-architecture pairs. This is an important observation, as it suggests that over-parameterized models are not only generalizable but also robust to adversarial samples. However, this outcome may be due to a failure of the PGD attack used for the evaluation.

\begin{figure*}[!htb]
    \centering
    \begin{subfigure}[b]{0.3\textwidth}
    \centering
    \includegraphics[width=\textwidth]{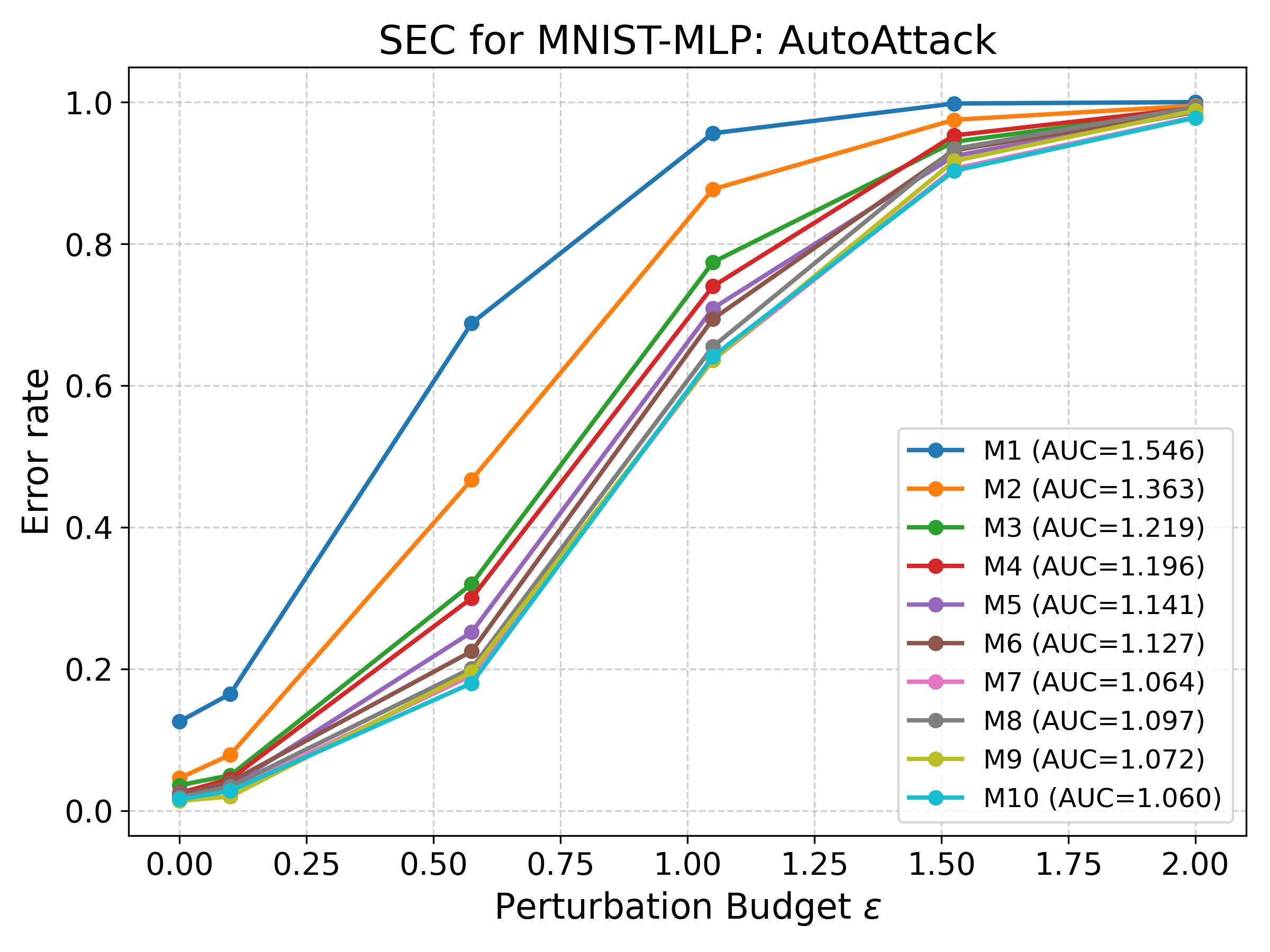}
    \caption{}
    \end{subfigure}
    \begin{subfigure}[b]{0.3\textwidth}
    \centering
    \includegraphics[width=\textwidth]{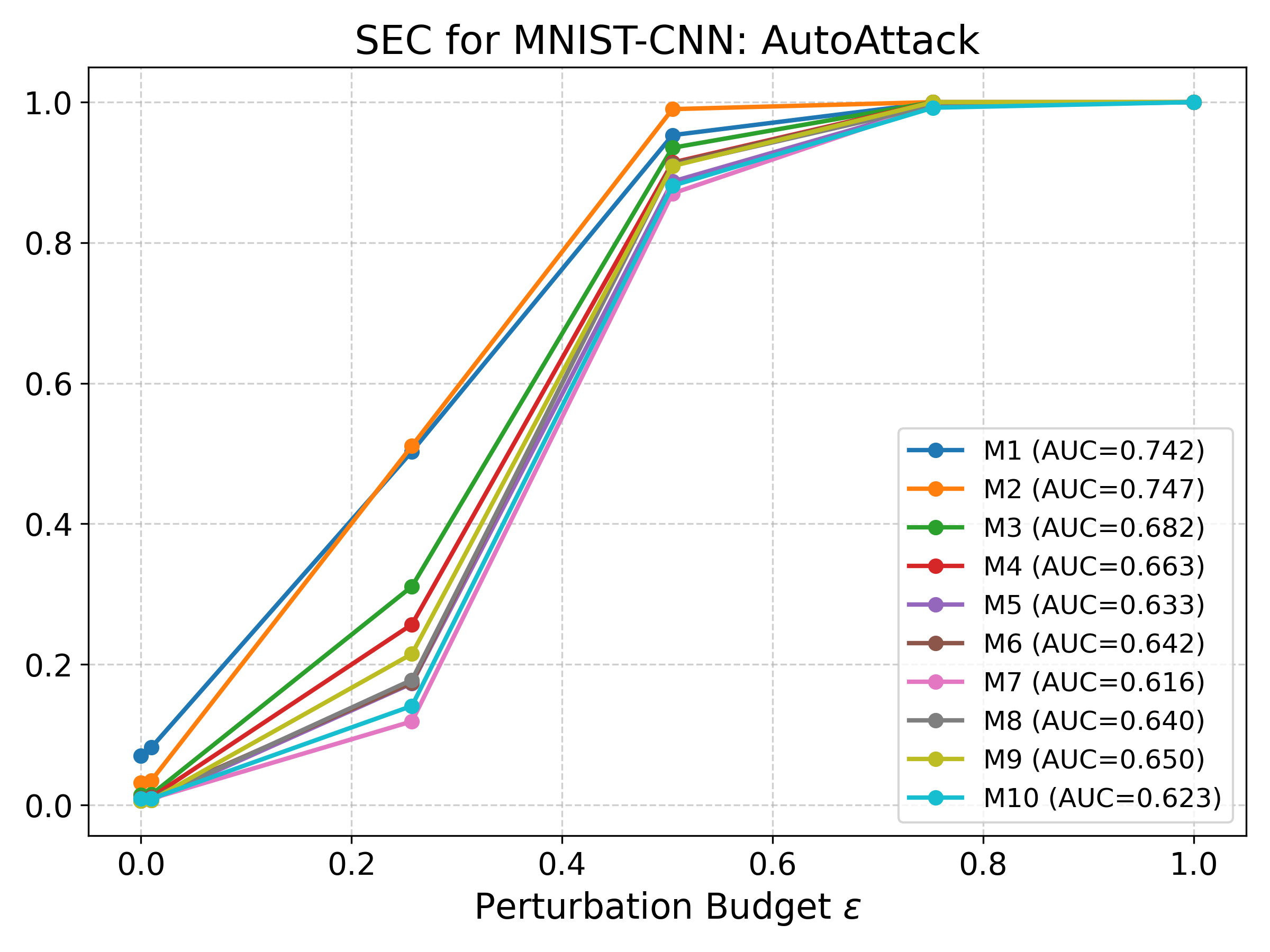}
    \caption{}
    \end{subfigure}
    \begin{subfigure}[b]{0.3\textwidth}
    \centering
    \includegraphics[width=\textwidth]{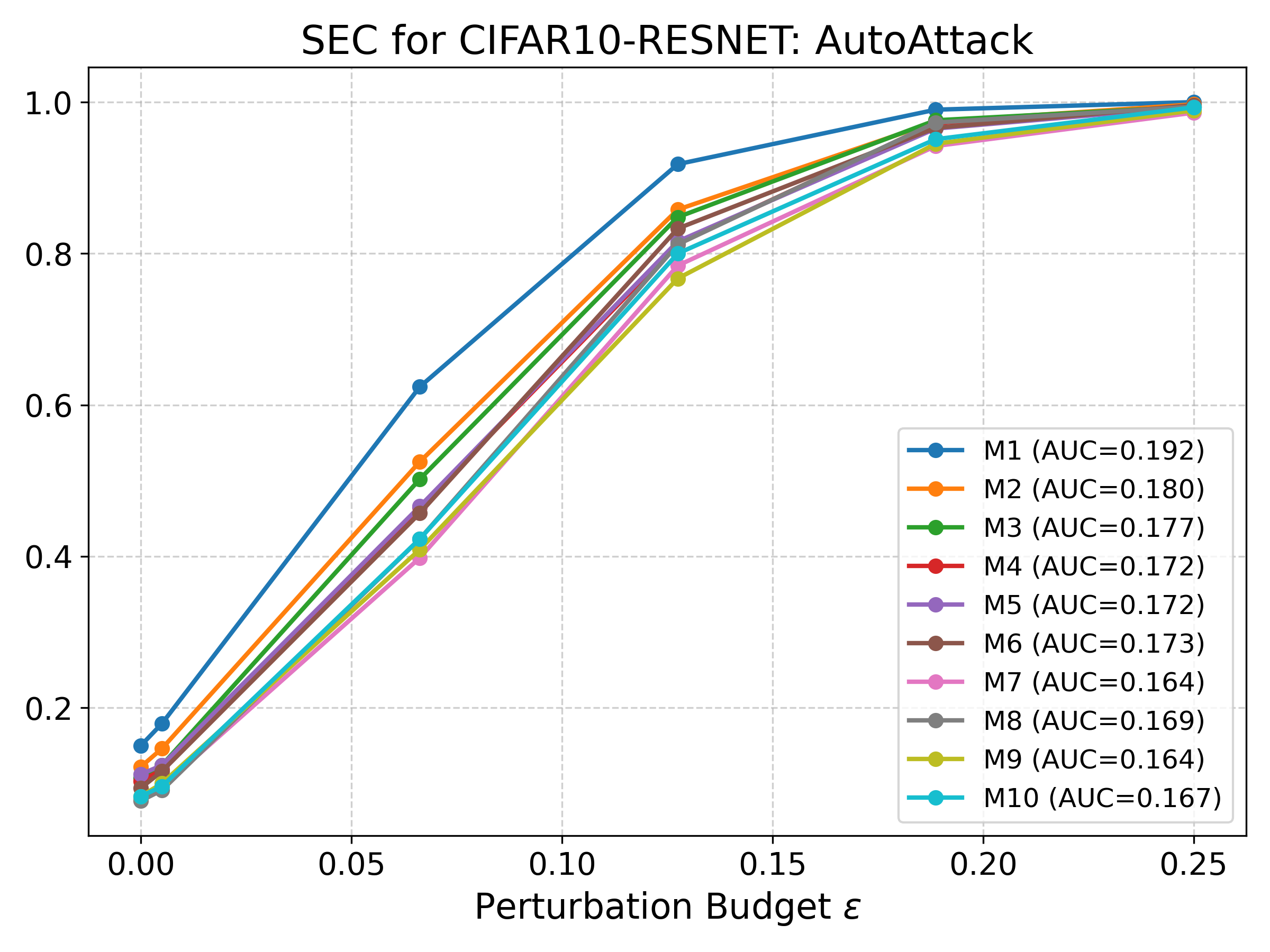}
    \caption{}
    \end{subfigure}
    \hfill
    \caption{\textbf{SEC Plots.} Evaluation of varying-sized models under \textbf{AutoAttack}: (a) MLP trained on MNIST, (b) CNN trained on MNIST, and (c) ResNet trained on CIFAR10. Each curve represents the security evaluation curve with the area under the curve specified in the legend.}
	\label{fig:performance_autoattack}
\end{figure*}

\subsubsection{Reliability of Model Robustness}

This subsection aims to verify the reliability of the analysis made with PGD.  

\paragraph{Evaluation of robustness with a standard benchmark}
The results in \autoref{fig:performance_autoattack} include all dataset-architecture configurations tested employing AutoAttack. 
Results:
\begin{itemize}
\item Even though one could train a close-to-perfect model on the unperturbed test set using smaller networks since there isn't a significant decline in the benign data performance, over-parameterized models significantly help with the robustness.
\item The trends obtained with AutoAttack follow that of PGD from \autoref{sec: model_robustness}; moreover, we have verified the affidability of the \autoref{sec: model_robustness} plots performing IoAF and we have observed no triggers for attack failures for any of the six indicators. These results verify that the analysis performed in this paper is conducted employing all the knowledge about the security of machine learning models available so far, providing reliability to the observed higher robustness of overparameterized networks.
\end{itemize}

\paragraph{Lipschitz Evaluation}
For completeness, we conduct a further robustness analysis using the Lipschitz constant as a metric. Recall that the Lipschitz constant of a neural network is directly related to its robustness: the certifiable robustness radius is inversely proportional to the Lipschitz constant, with the top-2 output margin serving as the proportionality numerator~\cite{hein2017formal}. Consequently, a larger Lipschitz constant generally yields a smaller certifiable robustness radius, modulated by the margin between the top-2 output components. From a generalization perspective, \citet{bubeck2021universal} established that a large Lipschitz constant is necessary for interpolating below the noise level in the non-overparameterized regime, linking smoothness to generalization under overparameterization.

To quantify this, we measure the Lipschitz constant of models across varying parameter counts with respect to the Euclidean norm. Given a composed model $f = f^{(L)} \circ \cdots \circ f^{(1)}$, we compute both an upper bound $C_u(f)$ and a lower bound estimate $C_l(f)$ of the Lipschitz constant. The upper bound exploits the sub-multiplicativity of the Lipschitz constant under composition:
\begin{equation}
    C_u(f) = \prod_{i} C_u(f^{(i)}),
\end{equation}
where the Lipschitz constant of convolutional layers is estimated via the singular value method of~\cite{sedghi2018singular}; residual layers of the form $f(x) = x + h(x)$ are bounded by $C_u(f) = 1 + C_u(h)$ via the triangle inequality; and for fully connected layers with weight matrix $W$, the Lipschitz constant equals the spectral norm $\|W\|_2$. Finally observe that \texttt{ReLU} activations are $1$-Lipschitz, being $\mathrm{Jac}(\texttt{ReLU})$ diagonal with only $0$ or $1$ values.

For the lower bound, we exploit the fact that the Lipschitz constant equals the supremum of the spectral norm of the Jacobian over the input space. This quantity is lower-bounded by sampling points $x$ from a dataset $\mathcal{D}$ and augmenting with Gaussian perturbations $\varepsilon \sim \mathcal{N}(0, \sigma)$, yielding
\begin{equation}
    C_l(f, x) = \max_{\varepsilon \sim \mathcal{N}(0,\sigma)}\, \max_{i \neq j}\, \|\nabla (f_i - f_j)(x + \varepsilon)\|_2,
\end{equation}
from which $C_l(f) = \max_{x \in \mathcal{D}}\, C_l(f, x)$. In our experiments, we evaluate over the full test set $\mathcal{D}$, drawing $100$ perturbation samples per input with $\sigma = 0.125$.

Results are reported in Appendix~\ref{sec:app_lipschitz}. Consistent with prior literature, the upper bound $C_u$ proves overly loose, exhibiting no meaningful decrease as model size increases. The lower bound $C_l$, however, aligns well with the robust accuracy results in \autoref{sec: model_robustness} for resnet and mlp models: where a higher robust accuracy aligned lower estimated Lipschitz constants, corroborating the theoretical connection between smoothness and robustness, and hence with over parameterization. 

\section{Conclusions and Future Works} \label{sec: results}
In this work, we empirically study how robust over-parameterized models are in adversarial environments as compared to their under-parameterized counterparts. These impressive generalizable properties of large models defy the widely accepted understanding of the model selection method that uses the Empirical Risk Minimization (ERM) principle, which is biased to select the smallest network that can fit the data well, and is often on the left side of the second descent of the double descent curve. From the classical statistical notion of bias-variance tradeoff, the optimal model is neither too large nor too small to avoid overfitting and underfitting, respectively. Yet the performance of large models outperforms that of smaller models, making their usage ubiquitous. However, when it comes to the security of the large models, the literature has contradictory claims in favor and against the robustness in an adversarial environment. This claims are made without verifying the reliability of the attacks and, for this reason, they are unreliable. Our results conducted with an attack of which we have verified that does not present any known failures, show that over-parameterized models not only generalize well but are also more robust to carefully crafted adversaries in their environment than their under-parameterized counterparts. We provide the empirical evidence to show that \textit{even when increasing the number of parameters is no more beneficial for the accuracy on clean data, it is beneficial for the robustness against adversarial examples}. Future directions of this work may include evaluating over-parameterized models against privacy attacks. It would also be interesting to investigate if it is possible to obtain smaller networks with accuracy and robustness comparable to those of over-parameterized networks to avoid computational burden.

\section{Expected Impact on Future Work}
While some work about the robustness of over-parameterized deep neural networks was present in the literature, their claim were contradictory, and their analysis were unreliable because the effectiveness of the performed attacks was not verified. The analysis presented in this work is verified with ad-hoc tools to analyze its reliability and shows that over-parameterized networks are more robust. We believe the empirical evidence presented in this manuscript serves as a starting point for theoretical work explaining why over-parametrized networks are more robust. Moreover, we believe that highlighting the reliability of robustness analysis for over-parameterized networks and the incoherence of the previously proposed analysis will lead to more reliable future work on this topic.  

\section{Funding}
This work has been carried out while Srishti Gupta was enrolled in the Italian National Doctorate on Artificial Intelligence run by Sapienza University of Rome in collaboration with the University of Cagliari. This work was also partially supported by project FAIR (PE00000013) under the NRRP MUR program funded by the EU – NGEU (CUP: J23C24000090007).

\section{Author contributions}
S.G. wrote the manuscript draft, performed the experiments, and made some updates to the code written by L.D. Z.C. made the initial exploratory experiments. L.D. wrote the code for the experiments currently reported in the paper. 
A.D. supervised Z.C. and S.G. 
B.B., X.F., and F.R. acquired funding. All authors reviewed the manuscript.

\section{Declarations}

\subsection{Data availability}
This manuscript does not report data generation and utilizes publicly available datasets. 

\subsection{Ethic Declaration} 
Not Applicable

\subsection{Consent to Participate}
Not Applicable

\subsection{Clinical Trial Number}
Not Applicable

\subsection{Consent to Publish}
Not Applicable

\subsection{Competing Interest}
The authors declare no competing interests.

\subsection{Declaration of Generative AI in Scientific Writing}
During the preparation of this work, the authors used Grammarly in order to edit and refine the language. After using this tool/service, the authors reviewed and edited the content as needed and take full responsibility for the content of the published article.

\bibliography{addref,bibDB}

\appendix

\section{Security Evaluation Curves for PGD $l_\infty$ Norm}
In this section, we support our claims by presenting the security evaluation curves for $L-\infty$ attacks on all three architectures: MLP, CNN, and ResNet, by scaling the size of the models.

\begin{figure*}[!htb]
    \centering
    \begin{subfigure}[b]{0.3\textwidth}
    \centering
    \includegraphics[width=\textwidth]{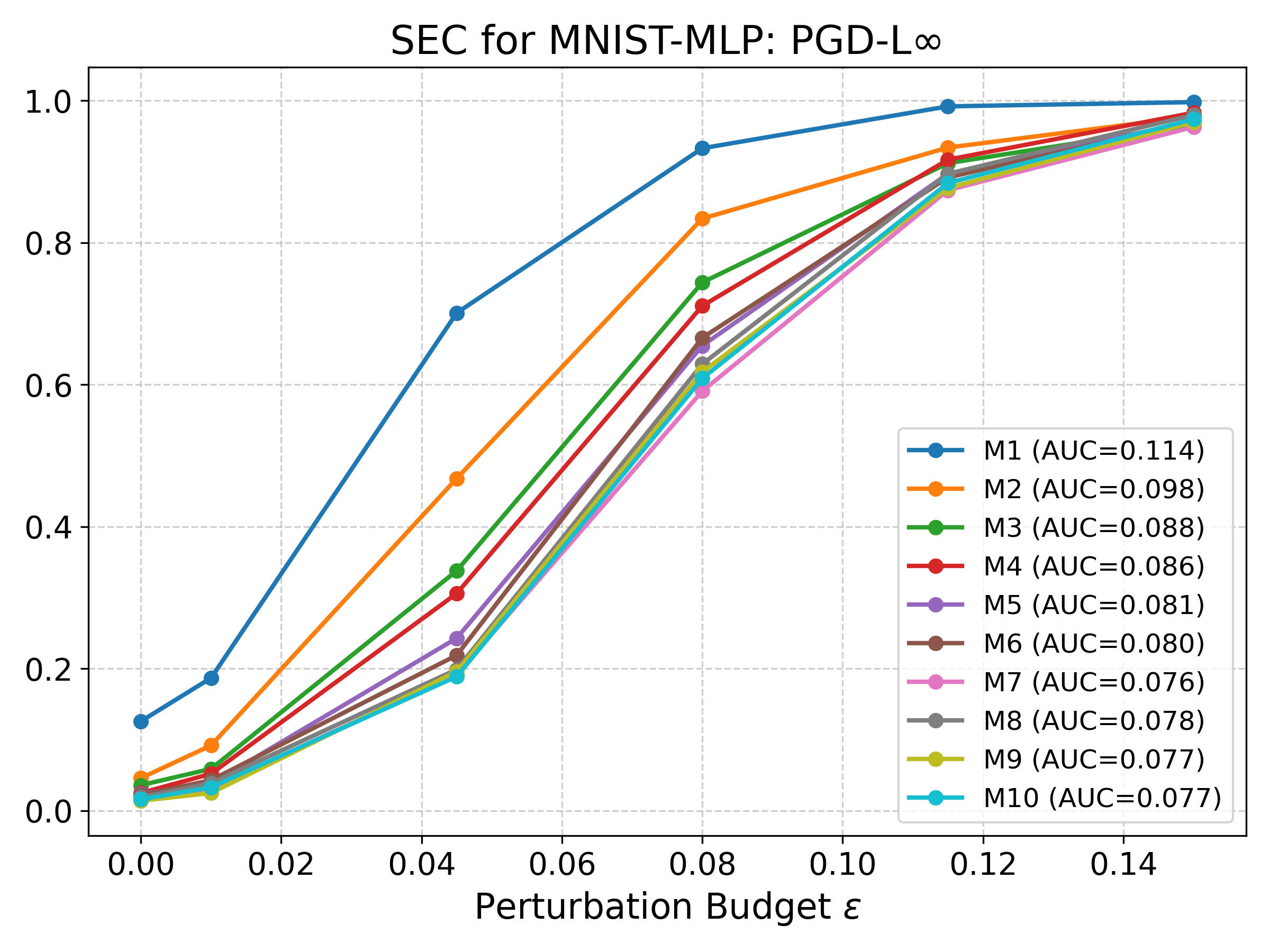}
    \caption{}
    \end{subfigure}
    \begin{subfigure}[b]{0.3\textwidth}
    \centering
    \includegraphics[width=\textwidth]{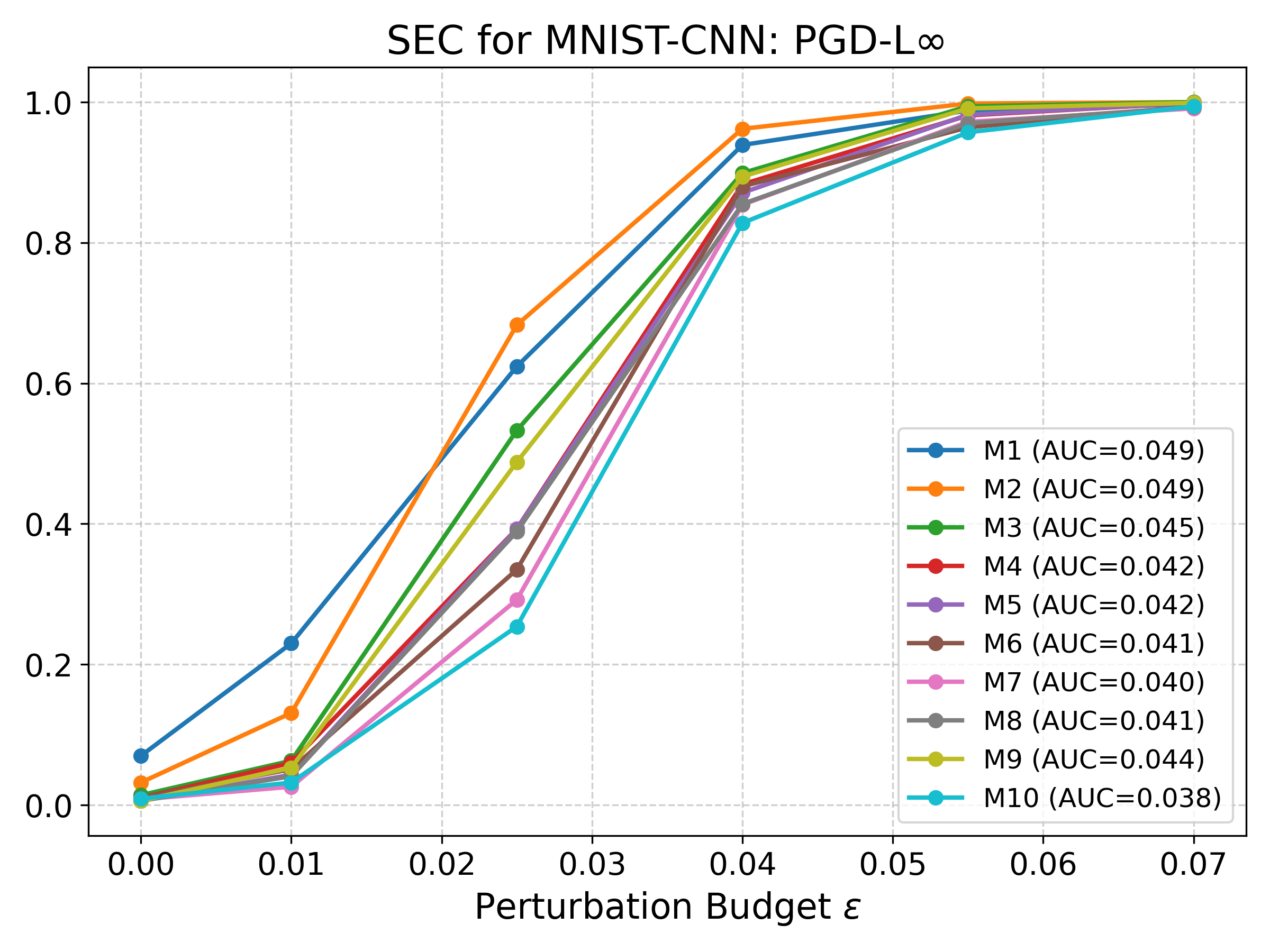}
    \caption{}
    \end{subfigure}
    \begin{subfigure}[b]{0.3\textwidth}
    \centering
    \includegraphics[width=\textwidth]{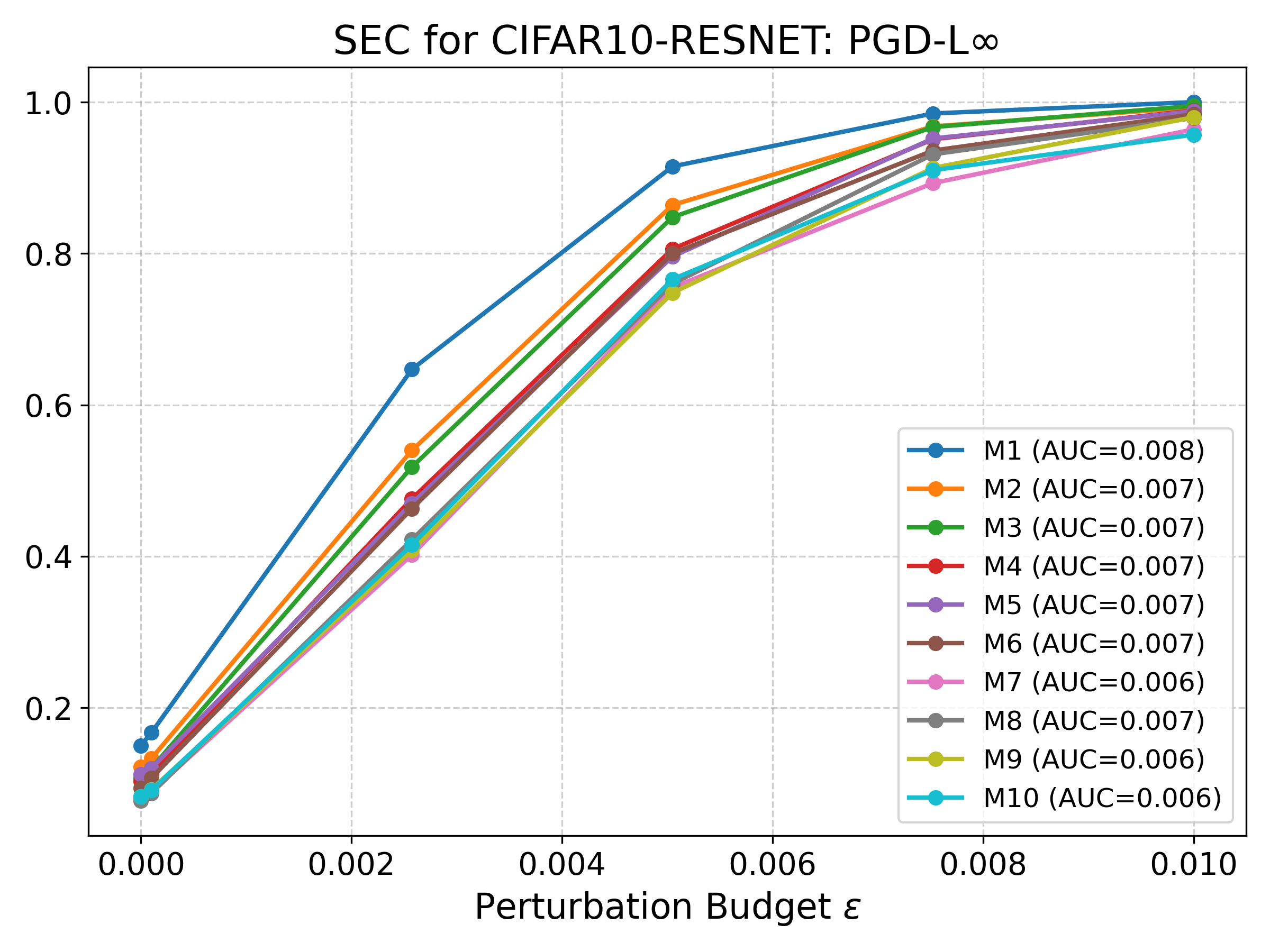}
    \caption{}
    \end{subfigure}
    \hfill
    
    \caption{\textbf{SEC Plots.} Evaluation of varying-sized models under \textbf{PGD $l_\infty$} attack: (a) MNIST trained on MLP, (b) MNIST trained on CNN, and (c) CIFAR10 trained on ResNet. Each curve represents the security evaluation curve with the area under the curve specified in the legend.}
    \label{fig:performance_pgdl_inf}
\end{figure*}

\section{Visualization of Adversarial Examples}

\begin{figure}
    \centering
    \includegraphics[width=0.75\linewidth]{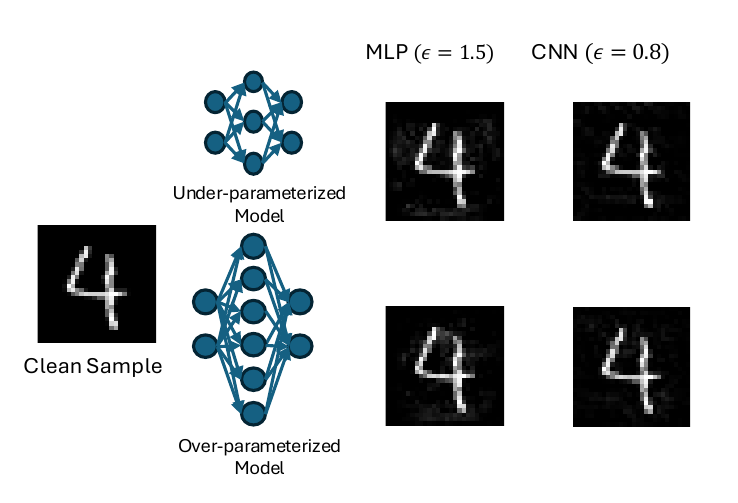}
    \caption{\textbf{Adversarial Examples.} The visual representation of PGD injected adversarial noise to MNIST digit 4 for under-parameterized (smallest) and over-parameterized (largest) models for corresponding architectures.: MLP($\epsilon=1.5$) and CNN ($\epsilon=0.8$). }
    \label{fig:visual-adv}
\end{figure}

In Figure~\autoref{fig:visual-adv}, we have visualized two examples of adversarial examples generated for an under-parameterized and an over-parameterized model for MLP and CNN.

\section{Bound Lipschitz }
\label{sec:app_lipschitz}
Overparameterization is a necessary but not sufficient condition for small Lipschitz interpolation.

\begin{figure*}[!htb]
    \centering
    \begin{subfigure}[b]{0.3\textwidth}
    \centering
    \includegraphics[width=\textwidth]{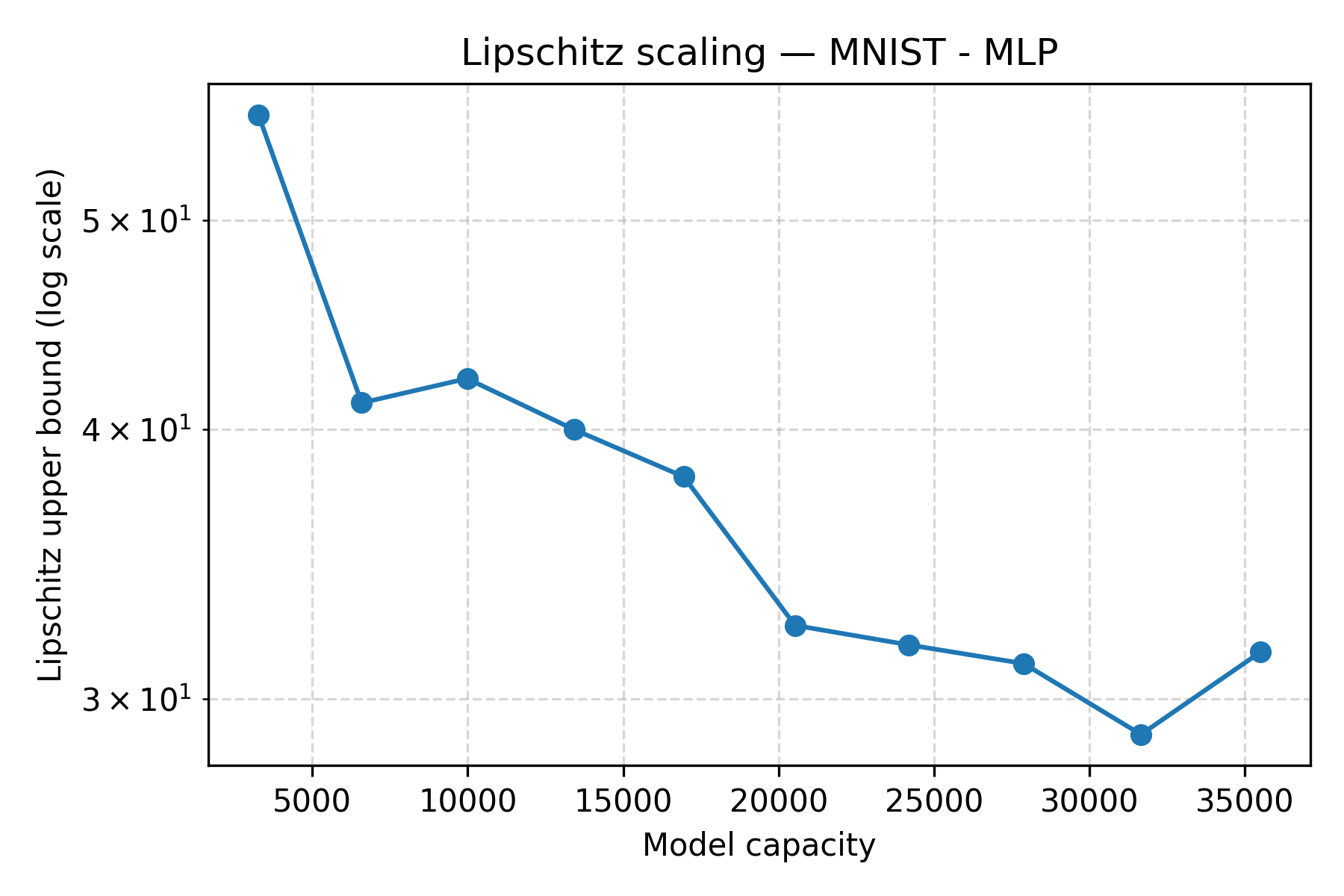}
    \caption{}
    \end{subfigure}
    \begin{subfigure}[b]{0.3\textwidth}
    \centering
    \includegraphics[width=\textwidth]{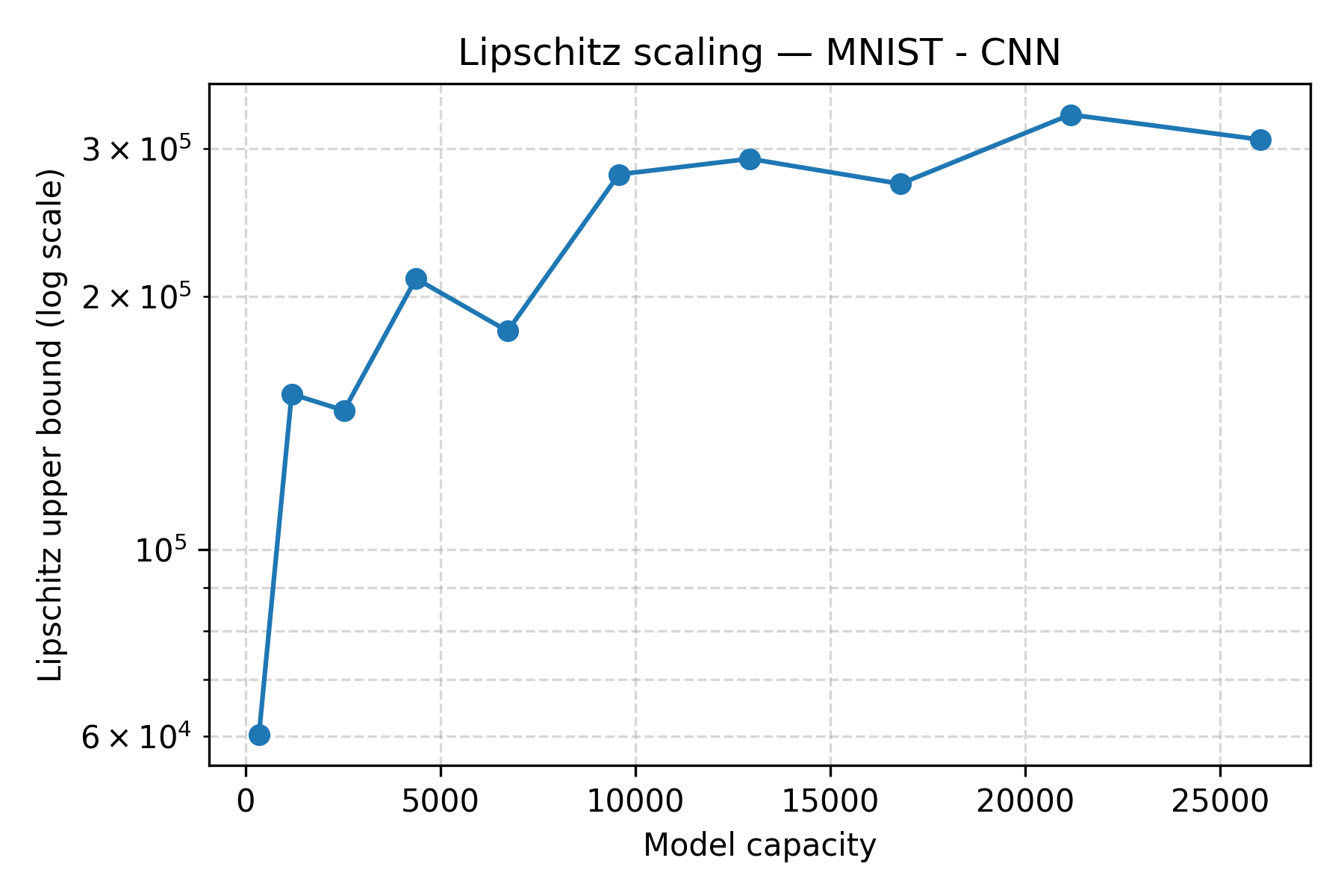}
    \caption{}
    \end{subfigure}
    \begin{subfigure}[b]{0.3\textwidth}
    \centering
    \includegraphics[width=\textwidth]{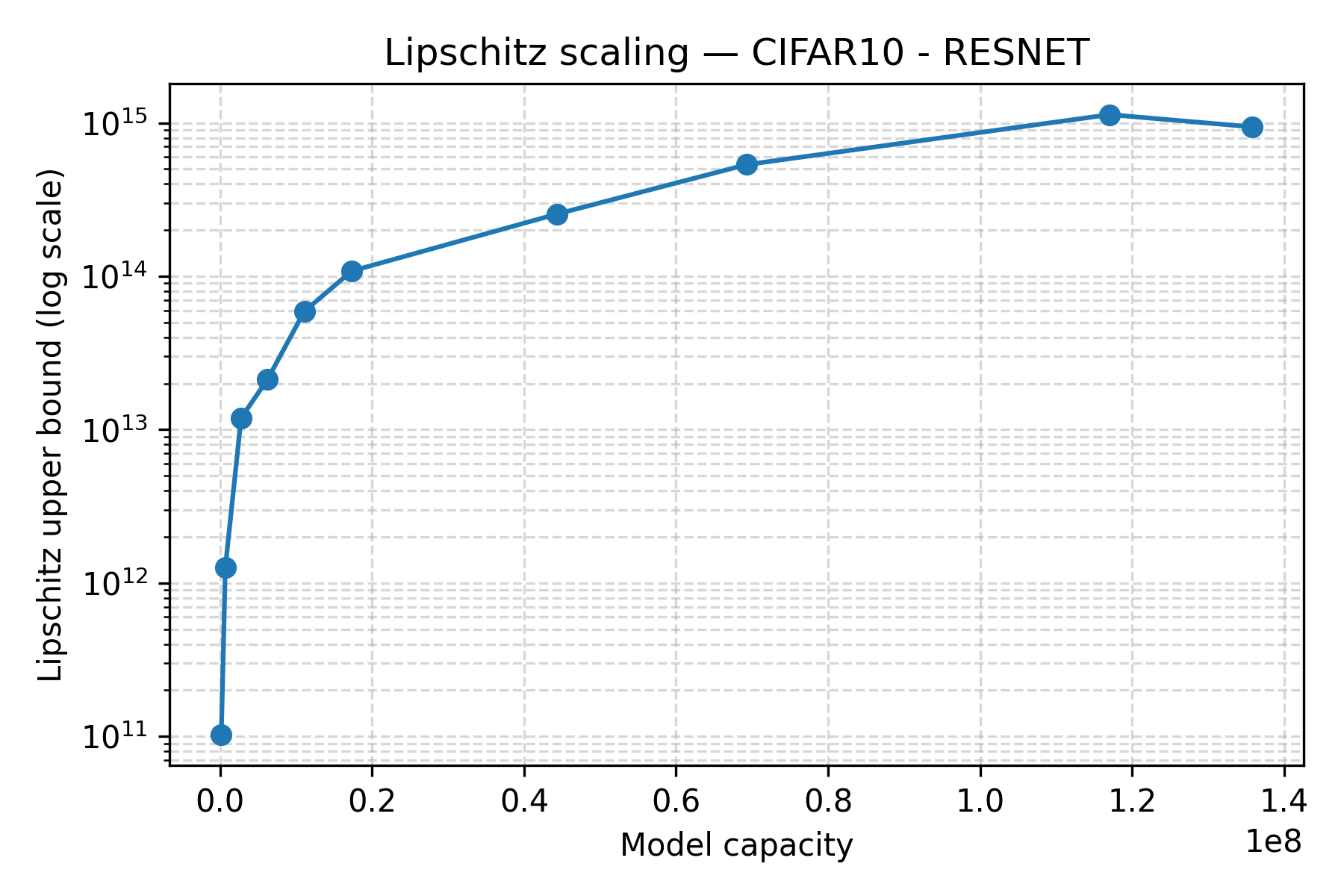}
    \caption{}
    \end{subfigure}
    \hfill
    
    \caption{\textbf{Upper Lipschitz Plots.} The upper bound of Lipschitz constant for: a) MLP, b) CNN, and c) ResNet. Each curve represents the Lipschitz value across different model sizes.}
    \label{fig:upper_lip}
\end{figure*}

\begin{figure*}[!htb]
    \centering
    \begin{subfigure}[b]{0.3\textwidth}
    \centering
    \includegraphics[width=\textwidth]{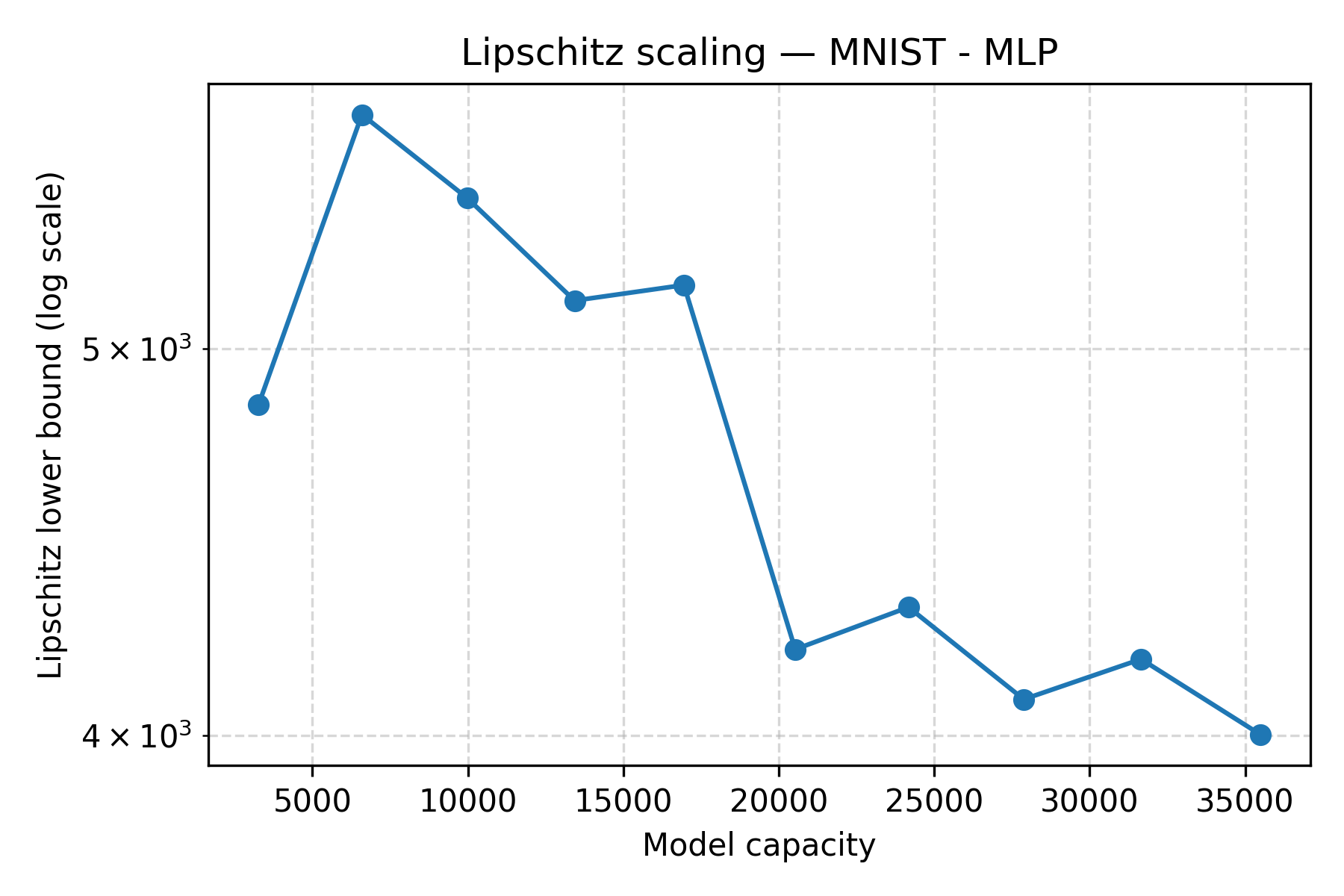}
    \caption{}
    \end{subfigure}
    \begin{subfigure}[b]{0.3\textwidth}
    \centering
    \includegraphics[width=\textwidth]{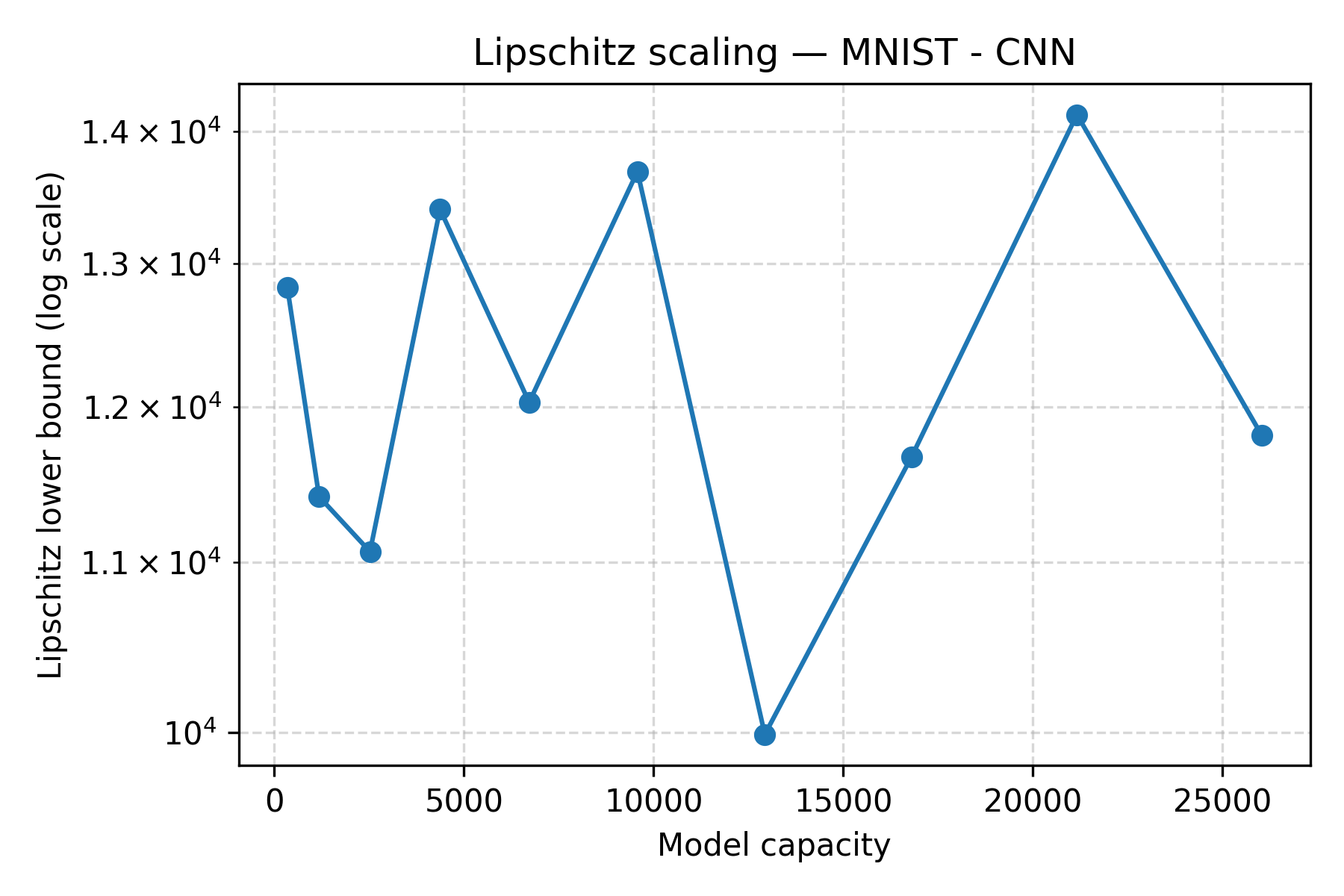}
    \caption{}
    \end{subfigure}
    \begin{subfigure}[b]{0.3\textwidth}
    \centering
    \includegraphics[width=\textwidth]{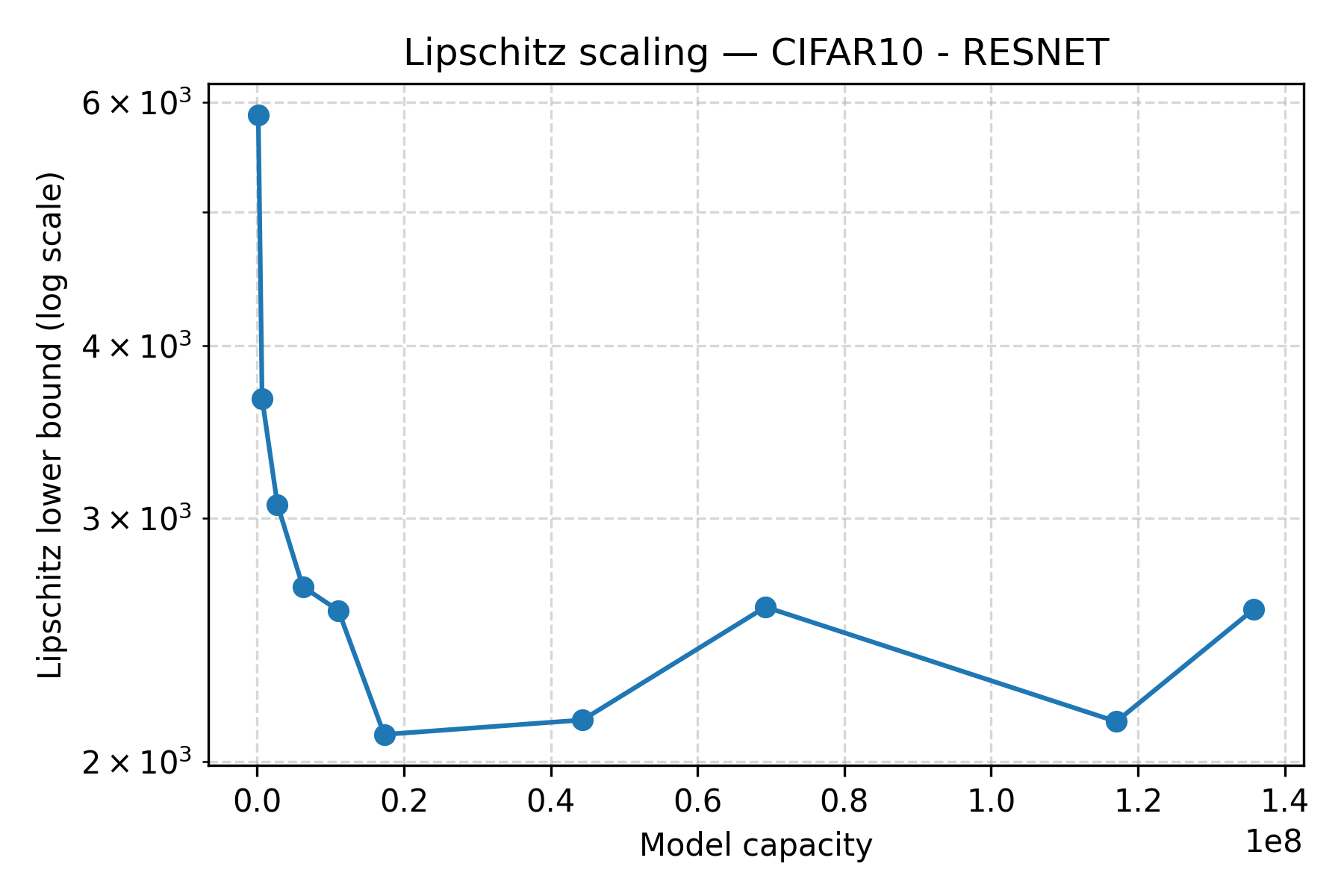}
    \caption{}
    \end{subfigure}
    \hfill
    
    \caption{\textbf{Lower Lipschitz Plots.} The lower bound of the Lipschitz constant for: a) MLP, b) CNN, and c) ResNet. Each curve represents the Lipschitz value across different model sizes.}
    \label{fig:lower_lip}
\end{figure*}

\end{document}